\documentclass[pmlr]{jmlr}

\RequirePackage{graphicx}
\usepackage{graphicx}
\usepackage{booktabs} 
\usepackage[utf8]{inputenc} 
\usepackage[T1]{fontenc}    
\usepackage{hyperref}       
\usepackage{url}            
\usepackage{amsfonts}       
\usepackage{nicefrac}       
\usepackage{microtype}      
\usepackage{amsmath}
\usepackage{multirow}
\usepackage{bm}
\usepackage{subcaption}
\usepackage{adjustbox}
\usepackage{enumitem}
\usepackage{soul}

\usepackage{hyperref}
\usepackage{longtable}
\makeatletter
\def\set@curr@file#1{\def\@curr@file{#1}} 
\makeatother
\usepackage[load-configurations=version-1]{siunitx} 

\theorembodyfont{\upshape}
\theoremheaderfont{\scshape}
\theorempostheader{:}
\theoremsep{\newline}

\jmlrvolume{VOLUME \# TBD}
\jmlryear{2025}

\title[Generating Accurate Synthetic Survival Data by Conditioning on Outcomes]{Generating Accurate Synthetic Survival Data by Conditioning on Outcomes}

\author{\Name{Mohammad Ashhad}
       \Email{mohammad.ashhad@kaust.edu.sa}\\ 
       \addr BESE\\
       King Abdullah University of Science and Technology (KAUST)\\
       Thuwal, KSA
       \AND
       \Name{Ricardo Henao}
       \Email{ricardo.henao@duke.edu}\\ 
       \addr Department of Bioinformatics \& Biotatistics\\
       Duke University\\
       Durham, USA} 

\begin{document}

\maketitle

\begin{abstract}
Synthetically generated data can improve privacy, fairness, and data accessibility; however, it can be challenging in specialized scenarios such as survival analysis.
One key challenge in this setting is censoring, {\em i.e.}, the timing of an event is unknown in some cases.
Existing methods struggle to accurately reproduce the distributions of both observed and censored event times when generating synthetic data.
We propose a conceptually simple approach that generates covariates conditioned on event times and censoring indicators by leveraging existing tabular data generation models without making assumptions about the mechanism underlying censoring.
Experiments on real-world datasets demonstrate that our method consistently outperforms baselines and improves downstream survival model performance.
\end{abstract}

\section{Introduction}

Synthetic data generation is the process of creating artificial data that mimic the statistical properties and patterns of real-world data.
This technique has gained significant attention in various machine learning settings, including data privacy and data augmentation \cite{Jordon2022}.
The primary motivation behind the generation of synthetic data is to address challenges associated with limited availability, privacy concerns, or population imbalance that is often prevalent in real-world data \citep{Zhang2017,Wang2021}.
For example, researchers and organizations could train and evaluate models using synthetic data without compromising sensitive or proprietary information.
Moreover, synthetic data can augment existing datasets, enabling more robust model performance.
Alternatively, it can protect data privacy by providing a means to share and exchange data without revealing sensitive information, facilitating collaboration across different domains \citep{Benedetti2020}.

Survival analysis, also known as time-to-event analysis, is a family of statistical methods that are used to analyze and model the time until the occurrence of a specific event (or outcome) of interest.
These methods are widely applied in various fields, including biomedical research, operations research, engineering, economics, and social sciences \citep{Kaso2022,Lillelund2023,Danacica2010,Gross2014}.
For example, assessing the effectiveness of medical treatments \citep{Singh2011}, predicting equipment failure rates \citep{CosJuez2010}, or analyzing customer churn in the business domain \citep{Danacica2010}.
The primary goal of survival analysis is to estimate the probability (distribution) of an event occurring over time, given a set of covariates or risk factors.
One of the distinctive challenges in survival analysis involves dealing with censored data, which occurs when the event of interest is not observed for some individuals within the study period.
This can happen for various reasons, such as loss of follow-up, measurement failure, study termination, or the occurrence of competing risks \citep{Salerno2023}.
The handling of censored data requires tailored statistical methods to avoid biased survival estimates.
Another challenge is that, oftentimes, sample sizes in survival data are relatively small, or the proportion of observed events relative to those with censoring is small, thus causing overfitting issues which negatively impact generalization ability.

\begin{figure*}[t]
    \includegraphics[width=\textwidth]{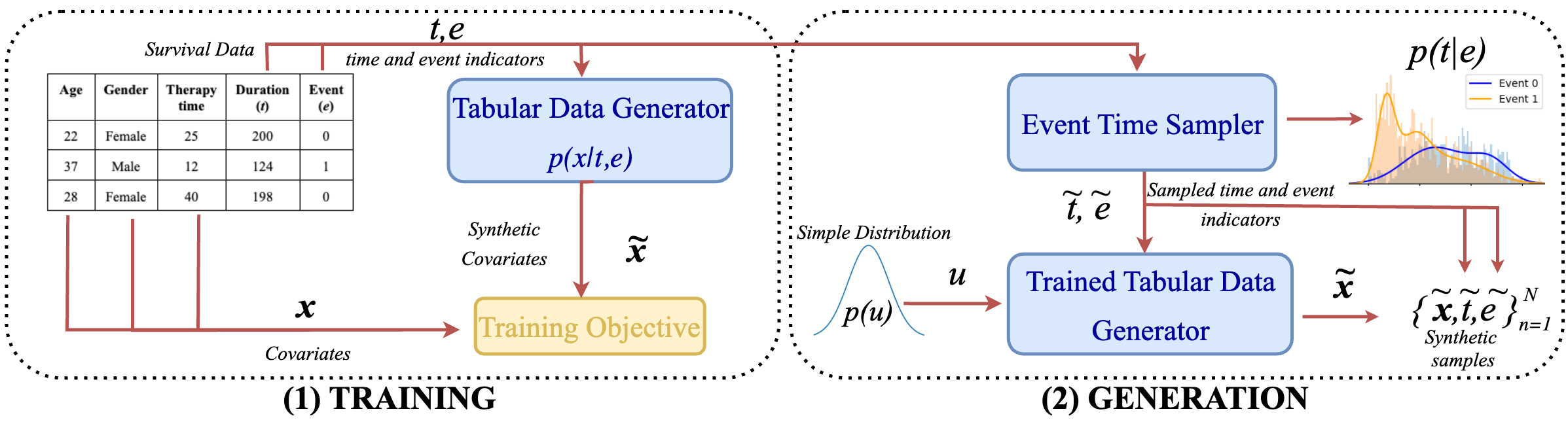}
    \caption{\small Block diagram of the proposed method.
    First, a conditional tabular data generator (we consider variational autoencoders, adversarial generators, diffusion-based models and large language models) is trained to learn to sample covariates from $p(\bm{x}|t,e)$.
    After training, times $\tilde{t}$ and event indicators $\tilde{e}$ are sampled from from a one-dimensional generator for $p(t,e=1)$ and $p(t,e=0)$ (we use a Dirichlet process mixture model, DPMM).
    These are then fed into the trained generator along with $\bm{u} \sim p(\bm{u})$, where $p(\bm{u})$ is a simple distribution. The generator then repeatedly generates synthetic covariates to completing the synthetic dataset ${\cal D}=\{(\bm{\tilde{x}},\tilde{t},\tilde{e})\}_{n=1}^N$.}
    \label{fig:architecture}
\end{figure*}

In most domains, such as clinical trials or engineering studies, collecting large amounts of survival data can be challenging, time-consuming, and costly.
Synthetic data generation allows researchers to create large datasets with desired characteristics, enabling more robust model prototyping, development, and evaluation.
Synthetic survival data, which are predominantly tabular (or structured), can be generated using generative models that are specifically developed for tabular data, {\em e.g.}, autoencoders \citep{Xu2019}, adversarial generators \citep{Yoon2020}, diffusion generators \citep{Kotelnikov2023}, and even large language models (LLMs) \citep{Borisov2022}.
However, in addition to the well-known challenges associated with generating tabular data, such as the appropriate handling of categorical and continuous data, mixed data types, and their joint distributions \citep{Xu2019}, survival data generation, especially in the medical domain, faces some unique challenges.
These are due to mainly unavoidable differences in the distributions for observed and censored events, and their (unknown) underlying generation mechanism given the covariates.
In practice, this challenge causes mismatches between these distributions when comparing real-world and synthetic data generated from it \citep{Norcliffe2023}.
Consequently, such mismatches are likely to cause survival models trained on such synthetic data to underperform relative to the real-world data in terms of discrimination and calibration.
So motivated, our work offers the following contributions:
\begin{itemize}[leftmargin=3mm,itemsep=0mm,topsep=0mm]
    \item We propose a simple method (see Figure~\ref{fig:architecture}) to generate survival data by conditioning the generation of covariates on event times and censoring indicators after sampling these from a model for the distribution of event times.
    \item We show that our {\em generator-agnostic} methodology can be easily extended to use LLM-based data generators to obtain high-quality synthetic survival data, an application that to the best of our knowledge has not been explored.
    \item Experiments on five real-world survival analysis datasets demonstrate the capabilities of the proposed method in terms of the quality of the generated observed, censored, and covariate distributions, as well as the discrimination and calibration performance of survival analysis models trained on synthetic data and evaluated on real-world data.
\end{itemize}

\subsection*{Generalizable Insights about Machine Learning in the Context of Healthcare}
\begin{itemize}[leftmargin=3mm,itemsep=0mm,topsep=0mm]
    \item \textbf{Synthetic data is crucial for advancing healthcare research while preserving privacy.} Healthcare datasets contain sensitive patient information, making them difficult to share across institutions. Synthetic data generation creates artificial data that mimics statistical properties of real clinical data without exposing patient identities, enabling broader collaboration while addressing limited data availability in specialized clinical studies.
    
    \item \textbf{Survival analysis poses unique challenges that require specialized modeling approaches.} The distinctive feature of survival data is censoring, where the timing of an event is unknown for some patients. Conventional data generation approaches struggle with correctly reproducing distributions of both observed and censored event times, which are critical for accurate prognostic assessment in healthcare.
    
    \item \textbf{Conditioning on event times and censoring indicators rather than covariates yields more accurate synthetic survival data.} Our approach of generating covariates conditioned on event times and censoring indicators preserves the critical time-to-event distributions by construction. This improves performance across multiple real-world clinical datasets while maintaining simplicity, making it practical for healthcare researchers.
\end{itemize}

\section{Related Work}\label{sc:related}

Generative models have emerged as powerful tools for synthesizing realistic data in various domains, including images, text, and tabular data.
These models aim to learn the underlying probability distributions of the training data and generate new samples that exhibit similar characteristics. 
Three prominent classes of generative models have gained significant traction: \emph{generative adversarial networks} (GANs), \emph{variational autoencoders} (VAEs), and \emph{diffusion-based models}.
GANs employ an adversarial training paradigm, where a generator network learns to produce synthetic data samples, while a discriminator network aims to distinguish between real and generated samples \citep{Goodfellow2020}.
This adversarial process drives the generator to produce increasingly realistic samples.
VAEs leverage variational inference techniques to learn a latent representation of the data, allowing the generation of new samples by sampling from the learned latent space \citep{Kingma2013}.
Diffusion-based models, such as \emph{denoising diffusion probabilistic model} (DDPM) \citep{Ho2020}, gradually add noise to the data and then learn to reverse this process, generating new samples by denoising random points.
These generative models have shown remarkable success in various applications, including image synthesis \citep{Kang2023}, text generation \citep{Su2022}, and video generation \citep{Jiang2023}.
{In survival analysis, generative models have been applied to estimate event-time distributions and hazard functions} \citep{Chapfuwa2018, Zhou2022}.

Tabular data stands out as a prevalent data format in machine learning (ML), with more than 65\% of datasets found on the Google Dataset Search platform (\url{datasetsearch.research.google.com}) comprising tabular files, typically in comma separated or spreadsheet formats \citep{Benjelloun2020}.
Although conventional generative methods are not optimally tailored for tabular data due to the mixture of continuous and categorical variables \citep{Xu2019}, modified versions have been developed for this domain.
These include the \emph{conditional tabular generative adversarial network} (CTGAN) \citep{Xu2019}, which leverages the GAN framework to generate synthetic data preserving multivariate distributions and relationships, the \emph{tabular variational autoencoder} (TVAE) \citep{Xu2019}, and the \emph{anonymization through data synthesis using generative adversarial network} (ADS-GAN) \citep{Yoon2020}.
The \emph{tabular denoising diffusion probabilistic model} (TabDDPM) is a recent approach that leverages denoising diffusion probabilistic models to generate high-fidelity synthetic tabular data \citep{Kotelnikov2023}.
{Large language models (LLMs) have also shown potential for tabular data generation, using fine-tuning on tabular data represented in token form} \citep{Borisov2022}.

In the generation of synthetic survival data, early statistical models \citep{Bender2005, Austin2012} transformed uniform samples into survival times but did not generate covariates.
More recent techniques have incorporated deep learning into the generative process.
\citet{Ranganath2016} proposed using deep exponential families to generate survival data, but this approach has limited flexibility on the learned distributions.
\cite{Miscouridou2018} and \cite{Zhou2022} relaxed this assumption but still focused on generating survival times and censoring statuses conditioned on the covariates, rather than generating the covariates themselves.
Recently, SurvivalGAN \citep{Norcliffe2023} was developed, generating synthetic data in three steps:
$i$) a conditional GAN (ADS-GAN) generates covariates ($\bm{x}$) and samples the event indicator ($e$) from the empirical distribution;
$ii$) a survival function model (DeepHit \citep{Lee2018}) predicts survival functions for the generated covariates; and
$iii$) these outputs are used by a regression model (XGboost \citep{Chen2016}) to predict the event-time ($t$), generating the complete triplet $(\bm{x},t,e)$. 
Although effective, this method is complex with multiple models, each having their own limitations.

Our work explores as an alternative a much simpler method that leverages existing generators for one-dimensional (event time) distributions and tabular (covariates) data and repurposes them for survival data without the need for dedicated networks for the prediction of the survival function or event/censoring distributions.

\section{Methods}

{\bf Problem Definition}
Instances (or subjects) of survival data can be represented in general as a triplet $z = (\bm{x}, t, e)$.
Here, $\bm{x} \in {\cal X}$ denotes $m$-dimensional tabular covariates that describe an instance's state at an initial (or index) time, encompassing both continuous and categorical covariates.
Then, $t_i \in {\cal T}$ represents the time of a specific event relative to the initial time, thus $t \ge 0$ and ${\cal T}\equiv\mathbb{R}_+$.
Lastly, $e_i \in {\cal E}$ stands for the event indicator, commonly ${\cal E} = \{0, 1\}$, where $e = 1$ indicates the event of interest occurs at time $t$, while $e = 0$ indicates that the event of interest has not occurred up to time $t$.
In this work, we only consider \emph{right censoring} as it is the predominant form in real-world datasets, however, the proposed method can be extended to left or interval censoring \cite{klein2006survival}.

{\bf Background}
Survival analysis is a statistical framework used to analyze and model the time until the occurrence of the event of interest, also known as the survival time or time-to-event.
Survival analysis involves modeling the conditional probability density function $p(t|\bm{x})$, to estimate the likelihood of the event of interest occurrs at time $t$ given the covariates $\bm{x}$.
From this, the survival function is derived, representing the probability that the event has not taken place by time $t$, {\em i.e.}, 
\begin{equation}\label{eq:surv}
      S(t \mid \boldsymbol{x})=\int_t^{\infty} p\left(t^{\prime} \mid \boldsymbol{x}\right) d t^{\prime} ,      
\end{equation}
where $S(t \mid \boldsymbol{x})$ is an estimate of the proportion of instances (subjects) with covariates $\bm{x}$ who have survived up to time $t$.
When the initial time is zero and given that events cannot occur at $t \leq 0$, thus $S(0|\bm{x}) = 1$.
Additionally, since $p(t|\bm{x})$ is a valid probability distribution (nonnegative), then $S(t|\bm{x})$ is a monotonically decreasing function.
Time-to-event approximation involves estimating the expected lifetime for any given covariate value, denoted as $\mu(\bm{x})$.
Specifically, this is obtained as $\mu(\bm{x}) = \int_{0}^{\infty} t^{\prime}p(t^{\prime}|\bm{x}) \, dt^{\prime}$, which, through integration by parts, simplifies to the area under the survival curve: $\mu(\bm{x}) = \int_{0}^{\infty} S(t|\bm{x}) \, dt$.

Survival models typically fall into one of two categories: $i$) parametric such as the accelerated failure time \citep{Weibull1951}, and log-logistic \citep{Prentice1976} models; or $ii$) non-parametric such as the Kaplan-Meier estimator \citep{Kaplan1958} and Cox proportional hazards model \citep{Cox1972}.
Moreover, deep-learning versions of these have been proposed, {\em e.g.}, DeepSurv \citep{Katzman2018}, DeepHit \citep{Lee2018}, DATE \citep{Chapfuwa2018}, {\em etc}.

{\bf Conditioning on Event Time and Type}
Synthetic survival data generation involves the generation of samples from the complete joint distribution $p(\bm{x},t,e)$.
In practice, one can either sample from it directly (and unconditionally) using generative models for tabular data, or via conditioning using for instance $p(t|\bm{x},e)p(\bm{x})p(e)$ or $p(\bm{x}|t,e)p(t|e)p(e)$.
The former is the approach used in \citet{Norcliffe2023}, in which the samples $\tilde{x}$ and $\tilde{e}$, from the marginals $p(\bm{x})$ and $p(e)$, are obtained using a conditional GAN (ADS-GAN) generator and the empirical distribution for the event indicators, respectively, and subsequently the samples $\tilde{t}$ from the conditional $p(t|\bm{x},e)$ are generated (deterministically) using a regression model.
One important drawback of this approach is that the quality of the samples for event times $\tilde{t}$ from $p(t|\bm{x},e)$ is both dependent on the quality of the approximation $\tilde{t} \sim p_\phi(t|\bm{x},e)$ (with parameters $\phi$) and that of $p(\bm{x})$ via $\tilde{\bm{x}} \sim p_\psi(\bm{x}|\bm{u})$ parameterized by $\psi$, and $\bm{u}$ being sampled from a simple distribution, {\em e.g.}, uniform or Gaussian.
As a result, approximation error in covariates $\bm{x}$ is compounded with that of $t$, resulting in event and censoring distributions that do not necessarily match the real data.
Consequently, \citet{Norcliffe2023} also proposed metrics to quantify the quality of these distributions (see Section \ref{sc:exp}).


In an effort to alleviate these key issues, we reverse the conditioning and instead sample \emph{both} event times and type from their joint distribution via $p(t|e)$ and $p(e)$, using a simple one-dimensional generator.
In our experiments, we consider a Dirichlet process mixture model (DPMM) \citep{Blei2006} to separately fit models for $p(t|e=0)$ and $p(t|e=1)$.
Note that this is possible by assuming without loss of generality that the observed and censoring times are conditionally independent given the covariates, which also aligns with the common assumption of censoring at random in survival analysis, which posits that the censoring mechanism is independent of the unobserved survival times, conditional on the covariates.
Then, we sample the covariates from $p(\bm{x}|t,e)$ using a conditional generator as follows:
\begin{align}\label{eq:xgivent}
    \tilde{e} \sim p(e), \hspace{1mm}
    \tilde{t} \sim p(t|\tilde{e}), \hspace{1mm}
    \bm{u} \sim p(\bm{u}), \hspace{1mm}
    \tilde{\bm{x}} \sim p_\theta(\bm{x}|\tilde{t},\tilde{e},\bm{u}),
\end{align}
where $p_\theta(\bm{x}|\tilde{t},\tilde{e},\bm{u})$ is a conditional generator parameterized by $\theta$, while $p(\bm{u})$ is a simple distribution.
Repeated sampling from the mechanism in \eqref{eq:xgivent} allows one to obtain a synthetic dataset ${\cal D}=\{(\bm{x}_n,t_n,e_n)\}_{n=1}^N$ whose empirical conditionals for event and censoring times readily match the ground-truth distributions, $p(t|e=1)$ and $p(t|e=0)$, respectively, and synthetic covariates that acknowledge their association with the event of interest while accounting for censoring.

Importantly, using \eqref{eq:xgivent}: $i$) eliminates the need for a supervised model to generate event times (XGboost in \citet{Norcliffe2023}); $ii$) eliminates the need for a separate model to generate survival distributions (DeepHit in \citet{Norcliffe2023}), and $iii$) guarantees the quality of the observed and censored event distributions.
Moreover, from a practical perspective, \eqref{eq:xgivent} offers flexibility, since $p_\theta(\bm{x}|\tilde{t},\tilde{e},\bm{u})$ can be modeled, in principle, with any conditional generator.
In the experiments, we consider TVAE, CTGAN, ADS-GAN, TabDDPM and LLMs.

Note that in \eqref{eq:xgivent} we are not required to sample from $p(t|e=0)$ and $p(t|e=1)$ using a DPMM.
Alternatively, one may use univariate (kernel) density estimators and then draw $\tilde{t}$ and $\tilde{e}$ accordingly, especially, if the dataset is small and the number of unique values of $t$ in ${\cal D}$ is small.
Moreover, in situations where privacy is a less sensitive concern, one could simply draw $p(t|e=0)$ and $p(t|e=1)$ from their empirical distributions.
We consider this setting in our experiments as a means to quantifying the impact on performance of using the DPMM to sample the event times.

\subsection{Adapting Conditional Generators to Survival Data}\label{sc:cond}
Existing tabular generators (see Section~\ref{sc:related}) use distinct strategies to implement conditioning.
Below, we briefly describe how they are adapted for survival data generation.

{\bf CTGAN}
This model being a conditional adversarial generator, synthesizes data using $G(\bm{u},\bm{c})$, where $G(\cdot)$ is the generator specified as a neural network, $\bm{u}$ is a vector sampled from a simple distribution, {\em e.g.}, a standard Gaussian distribution, {\em e.g.}, $\bm{u} \sim \mathcal{N}(\bm{0}, \bm{I})$, and $\bm{c}$ is a one-hot vector representing a discrete conditioning covariate.
See \cite{Xu2019} for additional details.
In order to use $G(\bm{u},\bm{c})$ as a sampling mechanism for $p_\theta(\bm{x}|\tilde{t},\tilde{e},\bm{u})$ in \eqref{eq:xgivent} we simply set $\bm{c}=E_t(\tilde{t})\oplus \tilde{e}$, where $E_t(\cdot)$ is an $m$-dimensional sinusoidal time embedding \citep{Wang2020} and $\oplus$ is the concatenation operator.
In our experiments, we set $m=4$.

{\bf TVAE}
The autoencoding formulation in \citet{Xu2019} does not specify explicitly how to perform conditional generation for the tabular VAE.
However, the simplest strategy involves setting the encoder and decoder pair as $\bm{u}\sim {\cal N}(\mu(\bm{x}),\sigma^2(\bm{x}))$ and $\tilde{\bm{x}}\sim p_\theta(\bm{x}|\bm{c},\bm{u})$, respectively, where here $\bm{u}$ is the latent representation for covariates $\bm{x}$, $\mu(\bm{x})$ and $\sigma^2(\bm{x})$ are two neural networks for the mean and variance functions of the latent representation $\bm{u}$, $p_\theta(\bm{x}|\bm{c},\bm{u})$ is a probabilistic decoder specified using neural networks (see \citet{Xu2019} for details), $\bm{c}$ is a one-hot vector as above for CTGAN, and the input to the decoder conveniently implemented by concatenating $\bm{z}$ and $\bm{c}$.
Similarly to CTGAN, we make $\bm{c}=E_t(\tilde{t})\oplus \tilde{e}$ in our implementation to sample from $p_\theta(\bm{x}|\tilde{t},\tilde{e},\bm{u})$ in \eqref{eq:xgivent} via $p_\theta(\bm{x}|\bm{c}=E_t(\tilde{t})\oplus\tilde{e},\bm{u})$.

{\bf ADS-GAN}
This alternative adversarial model specification encourages de-identifiability by letting the generator be $\tilde{\bm{x}}=G(\bm{u},\bm{x},\bm{c})$, {\em i.e.}, covariates $\bm{x}$ are also used as input to the generation function $G(\cdot)$, to encourage the model to generate samples $\tilde{\bm{x}}$ that are distinct from $\bm{x}$ to preserve privacy.
See \citet{Yoon2020} for additional details.
Consistent with CTGAN and TVAE above, we simply set $\bm{c}=E_t(\tilde{t})\oplus \tilde{e}$.

{\bf TabDDPM}
This model designed specifically for tabular data uses a combination of Gaussian and multinomial diffusion processes to handle numerical and categorical features, respectively.
Notably, each covariate uses a separate forward diffusion processes.
The reverse diffusion function in \citet{Kotelnikov2023} is set as $\bm{x}_{is}=g_i(\bm{x}_{i},\bm{x}_{i0},s)$, where $g_i(\cdot)$ is modeled using neural networks with identity and softmax activations for continuous and discrete covariates, respectively, $\bm{x}_{is}=h_x(\bm{x_{i}})+h_s(E_t(s))+E_c(\bm{c})$ is the representation of the $i$-th covariate in $\bm{x}$ at diffusion step $s$, $h_x(x_{i})$ is a fully connected layer with linear activation, $h_s(\cdot)$ is composed of two fully connected layers with sigmoid linear activations, $E_c(\cdot)$ is a standard (trainable) categorical embedding, and $s=0,\ldots,S$, is such that $\bm{x}_{iS}\sim{\cal N}(\bm{0},\bm{I})$ or $\bm{x}_{iS}\sim{\rm Cat}(\bm{1}/K_i)$, for $K_i$ categories (distinct values), for continuous or discrete covariates, respectively.
Note that, effectively, $g_i(\cdot)$ models the residuals of $\bm{x}_{is}$ at diffusion step $s$ rather than $\bm{x}_{is}$ itself \citep{Nichol2021}.
For additional details of the formulation and components of the model architecture, see \citet{Kotelnikov2023}.
For our implementation, we set $\bm{c}=E_t(\tilde{t})+E_s(\tilde{e})$ and set $m=128$ as the embedding dimension.

\section{Experiments}\label{sc:exp}
{\bf Baselines and setup}
We compare our methodology against the following baselines:
generative adversarial networks for anonymization (ADS-GAN) \citep{Yoon2020};
conditional generative adversarial networks for tabular data (CTGAN) \citep{Xu2019};
variational autoencoder for tabular data (TVAE) \citep{Xu2019};
tabular denoising diffusion probabilistic models (TabDDPM) \citep{Kotelnikov2023};
and SurvivalGAN \citep{Norcliffe2023}.
Note that only the latter is specific to survival data, whereas all the others generate tabular data {\em unconditionally}, {\em i.e.}, from the joint $p(\bm{x},t,e)$. 
For CTGAN, TVAE, ADS-GAN, and TabDDPM models, we report metrics both directly using the models {\em without} conditioning (Unconditional) for survival data generation, as well as our methodology (see Section~\ref{sc:cond}), {\em i.e.}, using them as conditional generators given event times and censoring indicators sampled using DPMMs.

To evaluate downstream performance, survival models are trained on synthetic data and tested on real data using the Train on Synthetic Test on Real (TSTR) paradigm \citep{Esteban2017}. {Specifically, the original dataset is divided into three folds, and the synthetic data generator is trained on two folds while the third is reserved for testing. Synthetic data equivalent (in size) to the training data is then generated, and downstream models are trained on this synthetic dataset and evaluated on the held-out {\em real test set}. This process is repeated for all three fold combinations.
We consider various survival models: linear (CoxPH)} \citep{Cox1972}, gradient boosting (SurvivalXGBoost) \citep{Barnwal2022}, and neural networks (DeepHit) \citep{Lee2018}, {and report metrics for the best-performing model. For each dataset, benchmark, and experimental setting, we report mean and standard deviation of performance metrics using 5 random seeds.}
To streamline the benchmarking, we utilized the Synthcity library \citep{Qian2024}, which provides implementations of a variety of synthetic tabular data generation models and benchmarking utilities.
Detailed experimental settings and hyperparameters are in Appendix~\ref{sc:appendix hyperparameters}. The source code for reproducing experiments is available at {\small\url{https://github.com/ashhadm/synthetic_survival_data}}.

{\bf Datasets}
We benchmark our methodology on a variety of real-world medical datasets.
Specifically:
$i$) \emph{Study to understand prognoses preferences outcomes and risks of treatment} (SUPPORT) \citep{Knaus1995};
$ii$) \emph{Molecular taxonomy of breast cancer international consortium} (METABRIC) \citep{Curtis2012};
$iii$) \emph{ACTG 320 clinical trial dataset} (AIDS) \citep{Hammer1997};
$iv$) \emph{Rotterdam \& German breast cancer study group} (GBSG) \citep{Schumacher1994}; and
$v$) \emph{Assay of serum free light chain} (FLCHAIN) \citep{Dispenzieri2012}. 
See Appendix~\ref{sc:appendix datasets} for additional details. 

{\bf Metrics}
To evaluate the quality of the generated synthetic survival data, various metrics are employed, which can be categorized into three groups:
{\em covariates quality}, {\em event-time distribution quality}, and {\em downstream performance}.
For assessing the quality of the generated covariates $\tilde{\bm{x}}$, the Jensen-Shannon (JS) distance and Wasserstein distance (WS) are used to measure the divergence between the generated and original covariate distributions.
For the quality of the event-time distributions we quantify the alignment between ground-truth and generated temporal marginals, namely, $p(t, e)$ is evaluated using the Kaplan-Meier (KM) divergence, optimism, and short-sightedness metrics as previously described in \citet{Norcliffe2023}.
The KM divergence compares the mean absolute difference between the synthetic and real survival function estimates, while optimism and short-sightedness are a proxy for their bias and variance, respectively. These three metrics capture the accuracy of the generated censoring and event distributions. Finally, to assess downstream performance, survival models are trained on the synthetic data and evaluated on real dataset. Specifically, we consider the concordance index (C-index) \citep{Harrell1982} and the Brier score \citep{Brier1950}. The former measures the discriminative ability of the survival model, while the latter quantifies the calibration of the probabilistic predictions.

\begin{table*}[t]
    \centering
    \caption{\small Quality, downstream and event-time metrics. Performance reported for best model between ADS-GAN, TVAE, CTGAN and TabDDPM. Original is for the survival model trained on the real (training) data. Subscripts are standard deviations for 5 repetitions.}
    \vspace{1mm}
    \adjustbox{width=0.95\textwidth}{
        \begin{tabular}{c c c c c c c}
        \textbf{Metric}&\textbf{Method} & \textbf{AIDS} & \textbf{METABRIC} & \textbf{SUPPORT} & \textbf{GBSG}& \textbf{FLCHAIN}\\
        \hline
         \multirow{4}{*}{JS distance ($\downarrow$)} & SurvivalGAN & 0.013$_{0.005}$ & 0.009$_{0.000}$ & 0.008$_{0.004}$ &0.008$_{0.001}$ &0.009$_{0.005}$\\
         &Unconditional & \textbf{0.006}$_{0.001}$ & 0.007$_{0.000}$ & 0.005$_{0.004}$ &0.005$_{0.002}$ &0.002$_{0.000}$ \\
         
        &Ours & \textbf{0.006}$_{0.005}$ & \textbf{0.006}$_{0.002}$ & \textbf{0.004}$_{0.005}$ &\textbf{0.004}$_{0.002}$ &\textbf{0.001}$_{0.003}$ \\
        
         \cmidrule{2-7}
        \multirow{4}{*}{WS distance ($\downarrow$)} & SurvivalGAN & 0.112$_{0.015}$ & 0.039$_{0.002}$ & 0.043$_{0.004}$ &0.019$_{0.005}$ &0.052$_{0.000}$\\
        &Unconditional & 0.065$_{0.005}$ & 0.031$_{0.005}$ & 0.036$_{0.005}$ &0.013$_{0.004}$ &\textbf{0.016}$_{0.005}$ \\
         
        &Ours& \textbf{0.063}$_{0.004}$ & \textbf{0.030}$_{0.002}$ & \textbf{0.032}$_{0.002}$ &\textbf{0.011}$_{0.004}$ &\textbf{0.016}$_{0.002}$ \\
         
       \hline
         
         \multirow{5}{*}{C-Index ($\uparrow$)} &\textcolor{gray}{Original} & \textcolor{gray}{0.760$_{0.005}$} & \textcolor{gray}{0.636$_{0.004}$} & \textcolor{gray}{0.616$_{0.002}$} & \textcolor{gray}{0.695$_{0.006}$} & \textcolor{gray}{0.870$_{0.004}$} \\
         & SurvivalGAN & 0.735$_{0.005}$ & 0.625$_{0.000}$ & 0.602$_{0.004}$ &0.668$_{0.005}$ &0.870$_{0.002}$\\
         
         &Unconditional & 0.779$_{0.002}$ & 0.649$_{0.004}$ & 0.625$_{0.002}$ &0.679$_{0.002}$ &0.879$_{0.004}$ \\
         
        &Ours & \textbf{0.785}$_{0.025}$ & \textbf{0.652}$_{0.005}$ & \textbf{0.626}$_{0.004}$ &\textbf{0.682}$_{0.002}$ &\textbf{0.880}$_{0.005}$ \\
        
        \cmidrule{2-7}
         
        \multirow{5}{*}{Brier Score ($\downarrow$)} &\textcolor{gray}{Original} & \textcolor{gray}{0.062$_{0.005}$} & \textcolor{gray}{0.200$_{0.004}$} & \textcolor{gray}{0.195$_{0.002}$} & \textcolor{gray}{0.205$_{0.005}$} & \textcolor{gray}{0.095$_{0.004}$} \\
        &  SurvivalGAN & 0.068$_{0.005}$ & 0.205$_{0.004}$ & 0.202$_{0.002}$ &0.212$_{0.005}$ &0.096$_{0.004}$\\
       &Unconditional & \textbf{0.060}$_{0.005}$ & 0.200$_{0.004}$ & 0.199$_{0.002}$ &\textbf{0.207}$_{0.005}$ &0.086$_{0.005}$ \\
        &Ours& \textbf{0.060}$_{0.004}$ & \textbf{0.197}$_{0.005}$ & \textbf{0.198}$_{0.002}$ &0.210$_{0.004}$ &\textbf{0.085}$_{0.004}$ \\
        
    \hline
         
       \multirow{4}{*}{Optimism ($\rightarrow 0$)} & SurvivalGAN & 0.021$_{0.005}$ & 0.011$_{0.002}$ & 0.016$_{0.004}$ & 0.006$_{0.003}$ & 0.134$_{0.005}$ \\
       &Unconditional & 0.002$_{0.002}$ & \textbf{0.001}$_{0.005}$ & \textbf{0.001}$_{0.003}$ &\textbf{0.004}$_{0.005}$ &0.005$_{0.004}$ \\
         
        &Ours & \textbf{0.001}$_{0.001}$ & \textbf{0.001}$_{0.001}$ & \textbf{-0.001}$_{0.001}$ &\textbf{-0.004}$_{0.000}$ &\textbf{-0.003}$_{0.001}$ \\
         \cmidrule{2-7}

        \multirow{4}{*}{Shortsightedness ($\rightarrow 0$)} & SurvivalGAN & 0.007$_{0.003}$ & 0.124$_{0.004}$ & 0.020$_{0.002}$ & 0.019$_{0.005}$ & 0.005$_{0.003}$ \\
      &Unconditional & 0.002$_{0.005}$ & \textbf{0.000}$_{0.002}$ & 0.002$_{0.003}$ &0.014$_{0.002}$ &0.003$_{0.005}$ \\
         
        &Ours & \textbf{0.001}$_{0.001}$ & \textbf{0.000}$_{0.001}$ & \textbf{0.000}$_{0.002}$ &\textbf{-0.012}$_{0.001}$ &\textbf{-0.002}$_{0.002}$ \\
        
         \cmidrule{2-7}
         
      \multirow{4}{*}{KM Divergence ($\downarrow$)} & SurvivalGAN & 0.021$_{0.004}$ & 0.082$_{0.005}$ & 0.064$_{0.002}$ & 0.049$_{0.003}$ & 0.134$_{0.004}$ \\
      &Unconditional & 0.015$_{0.003}$ & 0.019$_{0.003}$ & 0.011$_{0.004}$ &0.026$_{0.005}$ &0.007$_{0.002}$ \\
         
        &Ours & \textbf{0.003}$_{0.001}$ & \textbf{0.013}$_{0.002}$ & \textbf{0.006}$_{0.001}$ &\textbf{0.007}$_{0.001}$ &\textbf{0.005}$_{0.001}$ \\
         \bottomrule
         
        \end{tabular}
    }
    \label{tab:baselines}
\end{table*}

\subsection{Synthetic Survival Data Generation Benchmark}\label{sc: main exp}
\textbf{Covariate quality metrics:}
Results in Table~\ref{tab:baselines} compare the similarity between the distribution of synthetic samples and the original data.
First, we assess the overall (covariance) structure of the synthetic covariates relative to the original data via the JS and WS distances.
Our models outperformed or matched baselines in all 5 datasets for JS distance, and WS distance.
Full results are shown in Appendix \ref{sc:appendix metrics}.

\textbf{Downstream Performance} We conduct a comparative analysis of survival models trained with synthetic data generated by our methodology against models trained with data from baseline methods.
A favorable outcome is achieved when a model trained with synthetic data performs comparably to or occasionally even better than a model trained with real data, while also outperforming models trained with alternative synthetic data sources.
For reference, we also report the C-Index and Brier Score for survival models trained on the original data.
C-index and Brier score serve as the most widely used indicators of performance, as they encapsulate the entire conditional distribution of covariates, event/censoring times, and event indicators $p(t,e|x)$. Results in Table~\ref{tab:baselines} demonstrate that in C-index, we outperform the baselines across all datasets while in Brier Score, we outperform the baselines in 4 of 5 datasets.
Further, in most cases, we were also able to achieve better performance than survival models trained on the original data.

\textbf{Event-time distribution metrics} The proposed method demonstrates superior performance in preserving the underlying event-time distribution characteristics across all datasets as shown in Table~\ref{tab:baselines}.
Our approach achieves notably lower KM divergence scores compared to both SurvivalGAN and Unconditional baselines.
The method also exhibits minimal optimism and shortsightedness, with values consistently close to the ideal score of zero across all datasets.

\begin{table*}[h]
    \centering
    \small
    \caption{\small Quality and downstream performance metrics for synthetic survival data generation using LLMs. $Ours$ refer to our best model between ADS-GAN, TVAE, CTGAN and TabDDPM from Table \ref{tab:baselines}. Subscripts are standard deviations for 5 repetitions.}
    \vspace{1mm}
    \adjustbox{width=\textwidth}{
        \begin{tabular}{c c c c c c c}
        \textbf{Metric} & \textbf{Method} & \textbf{AIDS} & \textbf{METABRIC} & \textbf{SUPPORT} & \textbf{GBSG} & \textbf{FLCHAIN} \\
        \hline
        \multirow{4}{*}{JS Distance ($\downarrow$)} & SurvivalGAN & 0.013$_{0.005}$ & 0.009$_{0.000}$ & 0.008$_{0.004}$ &0.008$_{0.001}$ &0.009$_{0.005}$\\
        &Ours & 0.006$_{0.005}$ & 0.006$_{0.002}$ & 0.004$_{0.005}$ &\textbf{0.004}$_{0.002}$ &\textbf{0.001}$_{0.003}$ \\
        & LLM  & 0.004$_{0.001}$ & 0.006$_{0.001}$ & 0.003$_{0.000}$ & 0.007$_{0.001}$ & \textbf{0.001}$_{0.000}$ \\
        & LLM $(Ours)$ & \textbf{0.003}$_{0.001}$ & \textbf{0.005}$_{0.001}$ & \textbf{0.002}$_{0.000}$ & 0.006$_{0.001}$ & \textbf{0.001}$_{0.000}$ \\
        
        \cmidrule{2-7}
        
        \multirow{4}{*}{WS Distance ($\downarrow$)} & SurvivalGAN & 0.112$_{0.015}$ & 0.039$_{0.002}$ & 0.043$_{0.004}$ &0.019$_{0.005}$ &0.052$_{0.000}$\\
        &Ours & 0.063$_{0.004}$ & 0.030$_{0.002}$ & 0.032$_{0.002}$ &0.011$_{0.004}$ &0.016$_{0.002}$ \\
        & LLM  & 0.046$_{0.001}$ & \textbf{0.000}$_{0.000}$ & \textbf{0.020}$_{0.002}$ & 0.011$_{0.001}$ & 0.020$_{0.002}$ \\
        & LLM $(Ours)$ & \textbf{0.040}$_{0.003}$ & \textbf{0.000}$_{0.000}$ & 0.025$_{0.001}$ & \textbf{0.010}$_{0.001}$ & \textbf{0.015}$_{0.001}$ \\
        
        \midrule
        
        \multirow{4}{*}{C-Index ($\uparrow$)} & SurvivalGAN & 0.735$_{0.005}$ & 0.625$_{0.000}$ & 0.602$_{0.004}$ &0.668$_{0.005}$ &0.870$_{0.002}$\\
        &Ours & 0.785$_{0.025}$ & 0.652$_{0.005}$ & 0.626$_{0.004}$ &0.682$_{0.002}$ &\textbf{0.880}$_{0.005}$ \\
        & LLM  & 0.725$_{0.010}$ & 0.623$_{0.002}$ & 0.627$_{0.000}$ & 0.672$_{0.002}$ & 0.878$_{0.001}$ \\
        & LLM $(Ours)$ & \textbf{0.787}$_{0.002}$ & \textbf{0.655}$_{0.002}$ & \textbf{0.628}$_{0.001}$ & \textbf{0.684}$_{0.001}$ & \textbf{0.880}$_{0.000}$ \\
        
        \cmidrule{2-7}
        
        \multirow{4}{*}{Brier Score ($\downarrow$)} &  SurvivalGAN & 0.068$_{0.005}$ & 0.205$_{0.004}$ & 0.202$_{0.002}$ &0.212$_{0.005}$ &0.096$_{0.004}$\\
        &Ours & \textbf{0.060}$_{0.004}$ & 0.197$_{0.005}$ & \textbf{0.198}$_{0.002}$ &0.210$_{0.004}$ &0.085$_{0.004}$ \\
        & LLM  & 0.063$_{0.001}$ & 0.201$_{0.000}$ & 0.200$_{0.001}$ & \textbf{0.207}$_{0.001}$ & 0.090$_{0.001}$ \\
        & LLM $(Ours)$ & \textbf{0.060}$_{0.001}$ & \textbf{0.196}$_{0.001}$ & \textbf{0.198}$_{0.001}$ & \textbf{0.207}$_{0.002}$ & \textbf{0.083}$_{0.001}$ \\
        \bottomrule
        \end{tabular}
    }
    \label{tab:llm full}
\end{table*}

{\bf Ablation} In our ablation study, we investigated the impact of an alternate sampling strategy for time and event variables during the generation of synthetic data.
Although our base method uses DPMM to sample these variables, we also explore directly sampling \emph{both} event times and type from their joint distribution via $p(t|e)$ and $p(e)$, using their empirical distributions.
This modification led to modest but consistent improvements across all evaluation metrics while being remarkably simple and eliminating the need for a separate model to generate event times (See Table \ref{tab:ablation} in the Appendix).
Importantly, this performance gain did not come at the cost of privacy, as evidenced by the median and minimum Distance to Closest Record (DCR) values (see Table \ref{tab:privacy median full} and \ref{tab:privacy min full} in the Appendix).
In fact, the direct sampling approach achieved a higher median DCR in 3 out of 5 datasets and a higher minimum DCR in 4 out of 5 datasets when compared to \emph{unconditional} models, albeit by small margins. These results suggest that directly sampling $t$ and $e$ from the underlying distributions preserves the utility of the generated synthetic data while simplifying the generation process. Full results are shown in Appendix \ref{sc:appendix metrics}.
Based on the comprehensive evaluation metrics shown in the table, ADS-GAN consistently demonstrates superior performance in preserving the overall survival distribution, achieving the highest C-Index scores across most datasets while maintaining competitive Brier Scores. TVAE excels in terms of data fidelity metrics, showing particularly strong performance in JS and WS distances, making it particularly suitable when the priority is modeling the marginals of the original dataset. For overall utility and general-purpose synthetic survival data generation, ADS-GAN emerges as the most well-rounded model, offering the best balance between distribution preservation (high C-Index) and data fidelity (competitive JS/WS distances) while maintaining comparable performance in terms of calibration.

\subsection{Fine Tuning an LLM for Survival Data Generation}\label{sc:llm exp}
Generation of realistic tabular data (GReaT) is a recently proposed approach to generating high-quality synthetic tabular data using LLMs \cite{Borisov2022}. 
This is achieved by representing the tabular data as a sequence of text and training the language model to generate new sequences that correspond to valid and plausible tabular data instances.
We adapt GReaT to generate synthetic survival data by conditioning the generation on time-to-event and event-type.
The fine-tuning of a pre-trained auto-regressive LLM on the encoded tabular data for data generation as proposed in \citet{Borisov2022} involves the following steps. 

\begin{figure*}[t]
    \small
    \centering
    \includegraphics[width=\textwidth]{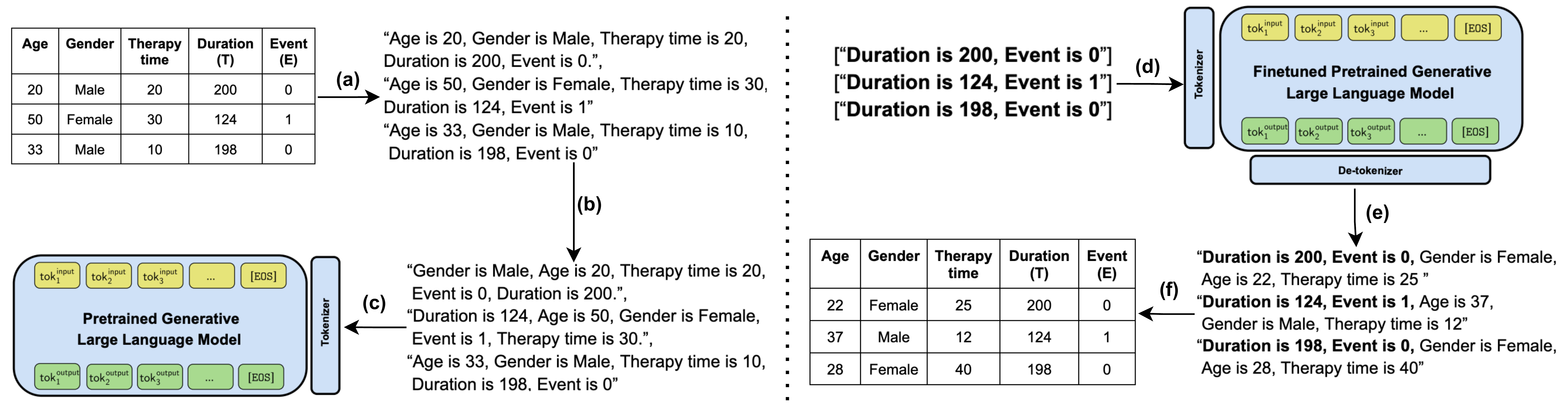}
    \caption{\small{Training and sampling procedure for survival data generation using LLMs.}}
\label{fig:llm}
\end{figure*}

{\em Textual encoding and feature permutation:} The tabular data with $M$ column names $\{f_m\}_{m=1}^M$ and thus, $M$-dimensional rows $\{\bm{x}_n\}_{n=1}^N$ are converted into textual representation.
Each row (sample) $\bm{x}_n$ is encoded as a sentence with elements $\bm{t}_n = \{t_{nm}\}_{m=1}^M$, where $t_{nm} = [f_m,``is", x_{nm}, ``,\ "]$ contains the column name $f_m$ and its value $x_{nm}$.
%
{\em Model training:} The LLM is trained using DistilGPT2 \citep{Li2021} on the textually encoded dataset $\{\bm{t}_{n}\}_{n=1}^N$, with elements of $\bm{t}_n$ permuted at random to remove pseudo-positional information as column order in a tabular dataset is in principle non-informative.
%
{\em Sampling:} Feature permutations during training enable the model to start generation with any combination of features and values.
To generate synthetic data conditionally, we prompt the trained model with conditioning sequences sampled from the DPMM model for $p(t,e)$, where survival times $\tilde{t}$ are drawn from the fitted DPMM for each event type $e \in {0,1}$, and let it generate the remaining tokens to complete the textual feature vector.
Unconditional generation follows \citet{Borisov2022}.
The training and sampling procedure is shown in Figure~\ref{fig:llm}.
Table~\ref{tab:llm full} compares the performance of GReaT with and without conditional generation (LLM $(Ours)$ and LLM respectively in Table \ref{tab:llm full}), against our best generator from Table \ref{tab:baselines}.
We observe that conditional generation consistently enhances LLM's performance over the unconditional variant and baseline generators.
Note however that GReaT is much more costly compared to other models as shown in Appendix~\ref{sc:appendix compute}. 

\subsection{Sub-population Level Evaluation of Synthetic Data}
In this experiment, we evaluate the performance of the proposed methodology at the sub-population level using the AIDS dataset, using race (White, Black and Hispanic) to define the sub-populations.
Performance evaluation is carried out via race-stratified $K$-fold cross-validation.
We consider survival models in three scenarios: $i$) trained on the real data; $ii)$ trained on synthetic data with the same race proportion as the original data (\emph{Synthetic}); and $iii$) trained on synthetic data with balanced race samples while preserving the distribution of observed and censored events for each race (\emph{Synthetic (Balanced)}).

Based on the results shown in Table~\ref{tab:exp 3}, for the survival models trained on the original AIDS dataset, the C-index differs across races, with the model performing better on Hispanic (0.778) when compared to White (0.724) and Black (0.724), with a $0.778/0.724\approx1.07$ ratio.
When training using our synthetic data (ADS-GAN conditioned on time and event) with the same distribution as the original data, the C-index values reflect a similar performance ratio of 1.06 between races.
For the balanced distribution scenario, all performance metrics improve at the expense of reducing the performance ratio between Hispanic and White/Black observed in the original data to 1.05.
Furthermore, the proposed model consistently outperforms SurvivalGAN, which is less able to capture the race performance difference with ratios 1.01 and 1.03 for Synthetic and Synthetic (Balanced), respectively.
\begin{table*}[t]
    \centering
    \caption{\small{Downstream performance metrics for survival models trained on Real Data, \emph{Synthetic}, and \emph{Synthetic (Balanced)}.}}
    \vspace{1mm}
    \adjustbox{width=0.8\textwidth}{
    \begin{tabular}{lcccccc}
        \multirow{2}{*}{\textbf{Method}} & \multirow{2}{*}{\textbf{Race}} & \multicolumn{2}{c}{\textbf{Synthetic}} & \multicolumn{2}{c}{\textbf{Synthetic (Balanced)}} \\
        \cmidrule(lr){3-4} \cmidrule(lr){5-6}
         & & \textbf{C-index} & \textbf{Brier Score} & \textbf{C-index} & \textbf{Brier Score} \\
        \midrule
        \multirow{4}{*}{ADS-GAN (Ours)}& All & 0.718$_{0.004}$ & 0.070$_{0.001}$ & 0.741$_{0.002}$ & 0.066$_{0.002}$ \\
                 & Race 1 & 0.718$_{0.002}$ & 0.067$_{0.002}$ & 0.723$_{0.001}$ & 0.062$_{0.002}$ \\
                 & Race 2 & 0.718$_{0.002}$ & 0.071$_{0.003}$ & 0.723$_{0.001}$ & 0.062$_{0.002}$ \\
                 & Race 3 & 0.760$_{0.005}$ & 0.071$_{0.003}$ & 0.761$_{0.010}$ & 0.062$_{0.001}$ \\
        \midrule
        
        \multirow{4}{*}{SurvivalGAN} & All & 0.663$_{0.004}$ & 0.100$_{0.020}$ & 0.683$_{0.010}$ & 0.076$_{0.010}$ \\
                     & Race 1 & 0.663$_{0.003}$ & 0.092$_{0.005}$ & 0.676$_{0.002}$ & 0.072$_{0.010}$ \\
                     & Race 2 & 0.663$_{0.003}$ & 0.095$_{0.010}$ & 0.676$_{0.010}$ & 0.073$_{0.020}$ \\
                     & Race 3 & 0.668$_{0.010}$ & 0.095$_{0.010}$ & 0.698$_{0.010}$ & 0.073$_{0.010}$ \\
        \midrule
        \multirow{2}{*}{\textbf{Method}} & \multirow{2}{*}{\textbf{Race}} & \multicolumn{4}{c}{\textbf{Real Data}} \\
        \cmidrule(lr){3-6}

         & &\multicolumn{2}{c}{\textbf{C-index}} & \multicolumn{2}{c}{\textbf{Brier Score}} \\
         \midrule
        \multirow{4}{*}{Original} &All &  \multicolumn{2}{c}{0.735$_{0.010}$} & \multicolumn{2}{c}{0.075$_{0.010}$} \\
        &Race 1 &  \multicolumn{2}{c}{0.724$_{0.001}$} & \multicolumn{2}{c}{0.069$_{0.002}$} \\
        &Race 2 &  \multicolumn{2}{c}{0.724$_{0.001}$} & \multicolumn{2}{c}{0.072$_{0.001}$} \\
        &Race 3 &  \multicolumn{2}{c}{0.778$_{0.020}$} & \multicolumn{2}{c}{0.072$_{0.001}$} \\
        \bottomrule
    \end{tabular}
    }
    \label{tab:exp 3}
\end{table*}

\subsection{Robustness to Limited Training Data}

To further verify the advantage of our reverse conditioning approach, we conducted an experiment examining how different methods perform with reduced training data. Using the AIDS dataset, we trained models on progressively smaller subsets (100\%, 75\%, and 50\%) of the original training data and evaluated their performance across different performance metrics. As shown in Table~\ref{tab:limited_data}, our reverse conditioning approach consistently outperforms both unconditional models and SurvivalGAN in all metrics, the performance gap widening as training data become more limited. With only 50\% of the training data, our method still maintains strong performance, while other approaches show a more substantial degradation.

\begin{table}[t]
\centering
\caption{Performance comparison with reduced training data on the AIDS dataset.}
\label{tab:limited_data}
\adjustbox{width=0.9\textwidth}{
\begin{tabular}{cccccc}
\textbf{Method} & \textbf{JS Distance} & \textbf{WS Distance} & \textbf{C-Index} & \textbf{Brier Score} & \textbf{KM Divergence} \\
\midrule
\multicolumn{6}{c}{\textbf{100\% Training Data}} \\
\midrule
Ours & \textbf{0.006$_{0.005}$} & \textbf{0.063$_{0.004}$} & \textbf{0.785$_{0.025}$} & \textbf{0.060$_{0.004}$} & \textbf{0.003$_{0.001}$} \\
Unconditional & \textbf{0.006$_{0.002}$} & 0.065$_{0.005}$ & 0.779$_{0.002}$ & \textbf{0.060$_{0.005}$} & 0.015$_{0.003}$ \\
SurvivalGAN & 0.013$_{0.005}$ & 0.112$_{0.015}$ & 0.735$_{0.005}$ & 0.068$_{0.005}$ & 0.021$_{0.004}$ \\
\midrule
\multicolumn{6}{c}{\textbf{75\% Training Data}} \\
\midrule
Ours & \textbf{0.007$_{0.004}$} & \textbf{0.069$_{0.005}$} & \textbf{0.780$_{0.023}$} & \textbf{0.062$_{0.005}$} & \textbf{0.004$_{0.001}$} \\
Unconditional & 0.008$_{0.003}$ & 0.072$_{0.006}$ & 0.770$_{0.015}$ & 0.063$_{0.005}$ & 0.018$_{0.004}$ \\
SurvivalGAN & 0.016$_{0.006}$ & 0.120$_{0.016}$ & 0.720$_{0.012}$ & 0.072$_{0.006}$ & 0.027$_{0.005}$ \\
\midrule
\multicolumn{6}{c}{\textbf{50\% Training Data}} \\
\midrule
Ours & \textbf{0.009$_{0.005}$} & \textbf{0.079$_{0.004}$} & \textbf{0.762$_{0.025}$} & \textbf{0.065$_{0.004}$} & \textbf{0.007$_{0.002}$} \\
Unconditional & 0.011$_{0.004}$ & 0.085$_{0.005}$ & 0.755$_{0.018}$ & 0.068$_{0.005}$ & 0.025$_{0.005}$ \\
SurvivalGAN & 0.022$_{0.007}$ & 0.135$_{0.020}$ & 0.695$_{0.015}$ & 0.078$_{0.008}$ & 0.038$_{0.006}$ \\
\bottomrule
\end{tabular}
}
\end{table}

Importantly, KM Divergence, which directly measures how well the models capture the survival distribution, remains consistently better with our approach, showing a 3.6x improvement over unconditional models and 5.4x improvement over SurvivalGAN when using only 50\% of the training data. Similarly, our method's C-Index decreases by only 2.9\%  when reducing training data by half, compared to a 5.4\% decrease for SurvivalGAN.

These results demonstrate that directly conditioning the generation of covariates on event-time and censoring indicators provides greater robustness when training data is limited, a crucial advantage in healthcare settings where large annotated datasets are often difficult to obtain. By separating the generation of event times from covariates, our approach utilizes the information available in small datasets more efficiently, making it particularly suitable for practical clinical applications.

\section{Discussion}\label{sc:conclusion}
This work proposed a simple yet effective methodology for generating high-quality synthetic survival data by conditioning the generation of covariates on event times and censoring indicators sampled from a one-dimensional distribution approximated with a DPMM.
Through extensive experiments on multiple real-world datasets, we demonstrated that our approach outperforms several competitive baselines across various evaluation metrics that assess the quality of the generated covariate distributions, alignment with the ground-truth event-time distributions, and the downstream performance of survival models trained on the synthetic data.
Moreover, we showcased the applicability of LLMs for survival data generation by fine-tuning them in a conditional manner on the textual representations of tabular data and how the proposed method preserves the sub-population-level performance characteristics of real-world data. Finally, we also demonstrated that our method performs better than baselines when there is less data available for training, as is common in survival studies.

{\bf Limitations}
Despite promising results, our work has limitations.
First, the quality of generated data depends on the representativeness and diversity of the original dataset used for training the generative models.
If the training data exhibit biases or lack variability, these will propagate to the synthetic data.
Second, while our approach ensures accurate reproduction of event-time and censoring distributions, it does not consider time-varying covariates, which may be relevant in certain applications.
Finally, further research is needed to address bias and equity in survival data.
Though we examine survival models' behavior trained on synthetic data at a sub-population level, we acknowledge that bias and equity are multifaceted challenges extending beyond this study.
These are exciting avenues for further research.

\section{Broader Impact}\label{sc:appendix broader impacts}
The ability to generate realistic synthetic survival datasets can have far-reaching impacts across various domains, especially in privacy-sensitive applications like healthcare and clinical research.
Synthetic data can enable model development, benchmarking, and collaboration while preserving patient confidentiality and complying with data protection regulations.
Furthermore, our methodology can potentially address the common challenge of limited data availability in survival analysis by augmenting existing datasets or creating entirely new synthetic datasets tailored to specific requirements. 
While synthetic survival data is specific to the domain to which it is applied, limiting the potential for misuse, it is important to acknowledge the possibility of reinforcing biases present in the training data, as is the case with any generative model.
Though we aim to understand the behavior of survival models trained on synthetic data across sub-populations, we recognize that addressing bias and ensuring equity are complex challenges that extend beyond the scope of this study.
Thus, it is crucial to exercise caution and implement appropriate safeguards to mitigate potential biases and promote fairness in the development and deployment of such models. We have no conflicts of interest to declare. All datasets used are publicly available and de-identified. Our work complies with relevant data protection regulations. We encourage users of our method to carefully consider the ethical implications and potential biases when applying it to sensitive healthcare data.

\bibliography{main}

\newpage
\appendix
\section{Ethics Statement}\label{sc:appendix ethics}
This study focuses on synthetic survival data generation, which has important ethical implications. While our method aims to preserve patient privacy by generating synthetic data, we acknowledge the potential risks of reinforcing biases present in the original datasets. We have made efforts to evaluate our approach across different sub-populations to assess fairness, but further work is needed to fully address bias and equity concerns in survival analysis. The synthetic data generated should not be used for real clinical decision-making without extensive validation. We have no conflicts of interest to declare. All datasets used are publicly available and de-identified. Our work complies with relevant data protection regulations. We encourage users of our method to carefully consider the ethical implications and potential biases when applying it to sensitive healthcare data.

\section{Experimental Details}\label{sc:appendix exp dets}
\subsection{Computational Cost}\label{sc:appendix compute}
All experiments, except for the LLM fine-tuning (see Section~\ref{sc:exp}), were conducted on Google Colab Pro using a T4 GPU.
For the LLM fine-tuning experiments, an NVIDIA A100 GPU was utilized on Colab. 
In Table~\ref{tab:time} we report the training time per iteration (TTPI) along with the time taken for synthetic data generation (GT) for all models used in Section~\ref{sc: main exp}, while the training and generation time for Section~\ref{sc:llm exp}) are reported in Table~\ref{tab:time llm}. 

\begin{table}[t]
\caption{\small Training time per-iteration (TTPI) and generation time (GT) for synthetic survival data generation (in seconds).}
\vspace{1mm}
\adjustbox{width=\textwidth}{
\begin{tabular}{c c c c c c c}
\toprule
\textbf{Metric}&\textbf{Method} & \textbf{AIDS} & \textbf{METABRIC} & \textbf{SUPPORT} & \textbf{GBSG}& \textbf{FLCHAIN}\\
\hline
\multirow{9}{*}{TTPI ($\downarrow$)} & SurvivalGAN & 0.178$_{0.004}$ & 0.260$_{0.003}$ & 1.234$_{0.013}$ & 0.283$_{0.005}$ & 1.024$_{0.012}$\\
& TVAE & 0.081$_{0.003}$ & 0.126$_{0.002}$ & 0.725$_{0.005}$ & 0.134$_{0.003}$ & 0.527$_{0.004}$ \\
& TabDDPM & 0.056$_{0.001}$ & \textbf{0.048}$_{0.002}$ & \textbf{0.201}$_{0.001}$ & \textbf{0.061}$_{0.003}$ & \textbf{0.183}$_{0.002}$  \\
& CTGAN & 0.174$_{0.002}$ & 0.242$_{0.004}$ & 1.231$_{0.012}$ & 0.226$_{0.003}$ & 0.891$_{0.011}$  \\
& ADS-GAN & 0.141$_{0.001}$ & 0.235$_{0.003}$ & 1.139$_{0.011}$ & 0.252$_{0.004}$ & 0.827$_{0.013}$  \\

\cmidrule{2-7}
& TVAE $(u)$ & 0.136$_{0.004}$ & 0.186$_{0.003}$ & 1.023$_{0.014}$ & 0.187$_{0.002}$ & 0.735$_{0.005}$ \\
     
 & TabDDPM $(u)$ & \textbf{0.046}$_{0.003}$ & 0.050$_{0.002}$ & 0.215$_{0.003}$ & 0.070$_{0.004}$ & 0.183$_{0.002}$  \\
    
    & CTGAN $(u)$ & 0.193$_{0.004}$ & 0.282$_{0.003}$ & 1.312$_{0.012}$ & 0.287$_{0.003}$ & 1.028$_{0.011}$  \\
    
     & ADS-GAN $(u)$ & 0.214$_{0.003}$ & 0.291$_{0.004}$ & 1.404$_{0.015}$ & 0.306$_{0.002}$ & 1.061$_{0.013}$  \\
     \midrule
    \multirow{9}{*}{GT ($\downarrow$)} & SurvivalGAN & 0.396$_{0.014}$ & 0.421$_{0.023}$ & 0.896$_{0.085}$ & 0.407$_{0.054}$ & 0.715$_{0.053}$\\
     & TVAE & 0.084$_{0.002}$ & 0.109$_{0.003}$ & 0.365$_{0.031}$ & 0.114$_{0.013}$ & 0.243$_{0.011}$ \\
    & TabDDPM & 11.870$_{0.135}$ & 9.430$_{0.227}$ & 49.611$_{0.284}$ & 17.121$_{0.107}$ & 35.116$_{0.195}$  \\
    & CTGAN & \textbf{0.085}$_{0.013}$ & 0.089$_{0.014}$ & 0.156$_{0.003}$ & 0.068$_{0.009}$ & 0.103$_{0.009}$   \\
     & ADS-GAN & 0.082$_{0.012}$ & \textbf{0.085}$_{0.003}$ & \textbf{0.149}$_{0.004}$ & \textbf{0.064}$_{0.012}$ & \textbf{0.101}$_{0.014}$ \\
     \cmidrule{2-7}
     & TVAE $(u)$ & 0.128$_{0.023}$ & 0.135$_{0.004}$ & 0.468$_{0.095}$ & 0.124$_{0.003}$ & 0.281$_{0.117}$  \\
    & TabDDPM $(u)$ & 11.785$_{0.367}$ & 9.466$_{0.337}$ & 50.017$_{0.817}$ & 18.085$_{0.567}$ & 34.937$_{0.857}$  \\
    & CTGAN $(u)$ & 0.079$_{0.003}$ & 0.087$_{0.004}$ & 0.192$_{0.035}$ & 0.073$_{0.004}$ & 0.124$_{0.117}$ \\
     & ADS-GAN $(u)$ & 0.089$_{0.004}$ & 0.098$_{0.013}$ & 0.212$_{0.045}$ & 0.085$_{0.013}$ & 0.111$_{0.003}$  \\
     \bottomrule
    \end{tabular}
}
\label{tab:time}
\end{table}

\begin{table}[t]
    \centering
    \caption{\small Training and generation time for synthetic survival data generation using LLMs (in seconds).}
    \vspace{1mm}
    \adjustbox{width=\textwidth}{
        \begin{tabular}{c c c c c c c}
        \toprule
        \textbf{Metric}&\textbf{Method} & \textbf{AIDS} & \textbf{METABRIC} & \textbf{SUPPORT} & \textbf{GBSG}& \textbf{FLCHAIN}\\
        \hline
         \multirow{1}{*}{TTPI} & LLM (Ours) & 5.182$_{0.113}$ & 9.584$_{0.195}$ & 49.756$_{0.004}$ &6.698$_{0.214}$ &23.367$_{0.015}$\\
         \midrule
        \multirow{2}{*}{GT} & LLM & 14.237$_{0.153}$ & 121.451$_{0.207}$ & 270.798$_{0.994}$ & 23.516$_{0.054}$ & 77.154$_{0.187}$\\
         & LLM (Ours) & 623.092$_{2.005}$ & 912.083$_{1.767}$ & 5519.845$_{5.574}$ & 812.298$_{0.256}$ & 1140.475$_{2.697}$ \\

         \bottomrule
         
        \end{tabular}
    }
    \label{tab:time llm}
\end{table}

\subsection{Datasets}\label{sc:appendix datasets}
We benchmark our methodology on a variety of medical datasets summarized in Table~\ref{tab:datasets}.
Specifically:
$i$) Study to understand prognoses preferences outcomes and risks of treatment (SUPPORT) \citep{Knaus1995};
$ii$) Molecular taxonomy of breast cancer international consortium (METABRIC) \citep{Curtis2012};
$iii$) ACTG 320 clinical trial dataset (AIDS) \citep{Hammer1997};
$iv$) Rotterdam \& German breast cancer study group (GBSG) \citep{Schumacher1994}; and
$v$) Assay of serum free light chain (FLCHAIN) \citep{Dispenzieri2012}.
Pre-processed versions of METABRIC, SUPPORT, and GBSG can be found at: \url{https://github.com/havakv/pycox}.
AIDS and FLCHAIN datasets can be downloaded from \url{https://github.com/sebp/scikit-survival/tree/master/sksurv/datasets/data}. 
For the FLCHAIN dataset, missing values in continuous covariates were imputed to the mean, while in discrete covariates they were imputed to the mode. 
All of these datasets are publicly available hence the experiments can be readily reproduced.
In parts of our code (see Section \ref{sc:cond} and \ref{sc:exp}), we utilize and modify the Synthcity library (\url{https://github.com/vanderschaarlab/synthcity}) which is protected under the \emph{Apache-2.0} license.
All rights to Synthcity are reserved by the original authors \citep{Qian2024}.

\begin{table}[t]
    \small
    \centering
    \caption{\small Summary statistics of the datasets used in the study.}
    \vspace{1mm}
    {
        \begin{tabular}{c|c|c|c}
        \hline
        \textbf{Dataset} & \textbf{No. instances} & \textbf{No. censored instances} & \textbf{No. features} \\
         \hline
         AIDS &  1151 & 96 & 11 \\
         METABRIC & 1904 & 801 & 9 \\
         FLCHAIN & 7874 & 5705 & 9   \\
        GBSG & 2232 & 965 & 7  \\
        SUPPORT & 8873 & 2837 & 14   \\
         \bottomrule

        \end{tabular}
    }
    \label{tab:datasets}
\end{table}

\clearpage

\subsection{Hyperparameters}\label{sc:appendix hyperparameters}
For reproducibility purposes, all hyperparameters are specified below.
Table~\ref{tab: hyperparameter survival} lists the hyperparameters for the downstream survival models used in the benchmarks.
Further, Tables~\ref{tab: hyperparameter llm} and \ref{tab:hyperparameters} provide the hyperparameters for all generative models employed in the study.

\begin{table}[h]
\small
\centering
\caption{\small Hyperparameters for the survival models used in Section \ref{sc:exp}.}
{
\vspace{1mm}
\begin{tabular}{lll}
\toprule
Method & Parameter & Parameter Value \\
\midrule
\multirow{3}{*}{CoxPH} & Estimation Method & Breslow \\
 & Penalizer & 0.0 \\
 & $L^1$ Ratio & 0.0 \\
\midrule
\multirow{8}{*}{SurvivalXGBoost} & Objective & Survival: AFT \\
 & Evaluation Metric & AFT Negative Log Likelihood \\
 & AFT Loss Distribution & Normal \\
 & AFT Loss Distribution Scale & 1.0 \\
 & No. Estimators & 100 \\
 & Column Subsample Ratio (by node) & 0.5 \\
 & Maximum Depth & 5 \\
 & Subsample Ratio & 0.5 \\
 & Learning Rate & $5 \times 10^{-2}$ \\
 & Minimum Child Weight & 50 \\
 & Tree Method & Histogram \\
 & Booster & Dart \\
\midrule
\multirow{8}{*}{Deephit} & No. Durations: & 1000 \\
 & Batch Size & 100 \\
 & Epochs & 2000 \\
 & Learning Rate & $1 \times 10^{-2}$ \\
 & Hidden Width & 300 \\
 & $\alpha$ & 0.28 \\
 & $\sigma$ & 0.38 \\
 & Dropout Rate & 0.2 \\
 & Patience & 20 \\
\bottomrule
\end{tabular}
}
\label{tab: hyperparameter survival}
\end{table}

\begin{table}[h]
\small
\centering
\caption{\small Hyperparameters used for the LLM in Section \ref{sc:llm exp}.}
\vspace{1mm}
{
\begin{tabular}{lll}
Method & Parameter & Parameter Value \\
\midrule
\multirow{6}{*}{GReaT (DistilGPT2)} & Batch Size & 32 \\
 & No. Iterations & 1000 \\
 & Learning Rate& $5 \times 10^{-5}$ \\
& Optimizer & AdamW \\
& Sampling Temperature& 0.7\\
& Sampling Batch Size&100\\
\bottomrule
\end{tabular}
}
\label{tab: hyperparameter llm}
\end{table}

\begin{table}[h]
\small
\centering
\caption{\small Hyperparameters of the generative models used in synthetic benchmarks in Section \ref{sc: main exp}.}
\adjustbox{width=0.8\textwidth}{
\begin{tabular}{lll}
\toprule
Model & Parameter & Parameter Value \\
\midrule
\multirow{13}{*}{ADS-GAN} & No. Iterations & 10000 \\
 & Generator no. Hidden Layers & 2 \\
 & Generator Hidden Units & 500 \\
 & Generator Non-linearity & ReLU \\
 & Generator Dropout Rate & 0.1 \\
 & Discriminator No. Hidden Layers & 2 \\
 & Discriminator Hidden Units & 500 \\
 & Discriminator Non-linearity & Leaky ReLU \\
 & Discriminator Dropout Rate & 0.1 \\
 & Learning Rate & $1 \times 10^{-3}$ \\
 & Weight Decay & $1 \times 10^{-3}$ \\
 & Batch Size & 200 \\
 & Gradient Penalty ($\lambda$) & 10 \\
 & Identifiability Penalty & 0.1 \\
 & Encoder Max Clusters & 5 \\
 & Early Stopping Patience & 5\\
\midrule
\multirow{12}{*}{CTGAN}  & No. Iterations & 2000 \\
 & Generator No. Hidden Layers & 2 \\
 & Generator Hidden Units & 500 \\
 & Generator Non-linearity & ReLU \\
 & Learning Rate & $1 \times 10^{-3}$ \\
 & Weight Decay & $1 \times 10^{-3}$ \\
 & Discriminator No. Hidden Layers & 2 \\
 & Discriminator Hidden Units & 500 \\
 & Discriminator Non-linearity & Leaky ReLU \\
 & Gradient Penalty ($\lambda$) & 10 \\
 & Batch Size & 200 \\
 & Early Stopping Patience & 5\\
 \midrule
 
 \multirow{4}{*}{SurvivalGAN} & Uncensoring Model & Survival Function Regression \\
 & Time-to-event strategy & Survival Function \\
 & Censoring Strategy & Random \\
 & Dataloader Sampling Strategy & Imbalance Time Censoring \\
 \midrule
\multirow{15}{*}{TVAE}  & No. Iterations & 1000 \\
 & Batch Size & 200 \\
& Learning Rate & $1 \times 10^{-3}$ \\
 & Weight Decay & $1 \times 10^{-5}$ \\
 & Encoder No. Hidden Layers & 3 \\
 & Encoder Hidden Units & 500 \\
 & Encoder Non-linearity & Leaky ReLU \\
 & Encoder Dropout Rate & 0.1 \\
 & Decoder No. Hidden Layers & 3 \\
 & Decoder Hidden Units & 500 \\
 & Decoder Non-linearity & Leaky ReLU \\
 & Decoder Dropout Rate & 0 \\
 & Early Stopping Patience & 5\\
 & Data Encoder Max Clusters & 10 \\
 & Embedding Width & 500\\

 \midrule
\multirow{7}{*}{TabDDPM}  & No. Iterations & 1000 \\
 & Batch Size & 1024 \\
& Learning Rate & $2 \times 10^{-3}$ \\
 & Weight Decay & $1 \times 10^{-4}$ \\
 & No. of Time-Steps & 1000 \\
 & Scheduler & Cosine \\
 & Gaussian Loss Type & MSE \\
\bottomrule
\end{tabular}
}
\label{tab:hyperparameters}
\end{table}

\clearpage
\section{Additional Performance Metrics}\label{sc:appendix metrics}

\textbf{Full quality metrics:} Below we provide the comprehensive scores of all models evaluated in the paper. Table~\ref{tab:full baselines} presents the {\em covariate quality} and {\em downstream performance} metrics for all models assessed in Section~\ref{sc: main exp}.
In Table~\ref{tab:event metrics}, we report the {\em event-time distribution quality} metrics, including optimism, short-sightedness, and KM Divergence, for both conditional and unconditional models. Although our base method uses DPMM to sample $t$ and $e$, we also explore directly sampling \emph{both} event times and type from their joint distribution via $p(t|e)$ and $p(e)$, using their empirical distributions. This comparison is shown in Table \ref{tab:ablation}.

\textbf{Privacy experiment:} To explore the acceptability of bootstrapping $t$ and $e$ when generating synthetic data, we employed the Distance to Closest Record (DCR) metric to evaluate the privacy preservation capabilities of various synthetic data generation methods \citep{Zhao2021}. The DCR quantifies the Euclidean distance between each synthetic record and its nearest real counterpart. A higher DCR value indicates a lower risk of privacy breach. We report the median and minimum DCR for all synthetic survival data generators used in our study, with the addition of a Synthetic Minority Oversampling Technique (SMOTE) \citep{Chawla2002} baseline. SMOTE, originally proposed for minority class oversampling, is a simple interpolation-based method that generates synthetic points as convex combinations of real data points and their $k$-th nearest neighbors. In this study, we generalized and applied SMOTE to synthetic data generation to bootstrap the entire data point $(\tilde{x},\tilde{t},\tilde{e})$, for comparison purposes. The results, presented in Table \ref{tab:privacy median full}, {demonstrate that the median DCR for the methods where $t$ and $e$ were bootstrapped (denoted by a dagger $\dagger$) was higher in 3 of 5 data sets, although by a small margin when compared to $unconditional$ models (denoted by $(u)$). A similar observation can be made in Table \ref{tab:privacy min full} {where minimum DCR was higher for our methods in 4 of 5 datasets compared to the $unconditional$ models. In general, the median and minimum DCR values were largely similar between the methods when sampling from empirical $t$ and $e$, sampling $t$ and $e$ from fitted DPMM or generating them along with the covariates as a joint distribution (unconditional) suggesting that sampling them is not likely to impact privacy. However, SMOTE consistently exhibited the lowest DCR in all datasets, indicating potential privacy concerns. These findings provide empirical evidence that bootstrapping $t$ and $e$ is generally acceptable from a privacy perspective.
However, we note that even the most stringent minimum DCR does not provide privacy guarantees, so it needs to be interpreted with care.

\begin{table}[h]
    \centering
    \small
    \caption{\small Quality and downstream performance metrics. Models conditioned on empirical $t$ and $e$ are highlighted ($\dagger$) and $u$ refers to unconditional models. Subscripts are standard deviations for 5 repetitions.}
    \vspace{1mm}
    \adjustbox{width=0.90\textwidth}{
        \begin{tabular}{c c c c c c c}
        \toprule
        \textbf{Metric}&\textbf{Method} & \textbf{AIDS} & \textbf{METABRIC} & \textbf{SUPPORT} & \textbf{GBSG}& \textbf{FLCHAIN}\\
        \hline
        
         \multirow{13}{*}{JS distance ($\downarrow$)} & SurvivalGAN & 0.013$_{0.005}$ & 0.009$_{0.000}$ & 0.008$_{0.004}$ & 0.008$_{0.001}$ & 0.009$_{0.005}$\\
         & TVAE$^\dagger$ & 0.007$_{0.004}$ & 0.008$_{0.005}$ & \textbf{0.004}$_{0.000}$ & 0.005$_{0.001}$ & 0.002$_{0.005}$ \\
        &TabDDPM$^\dagger$ & 0.007$_{0.000}$ & 0.007$_{0.001}$ & 0.013$_{0.005}$ & 0.005$_{0.002}$ & \textbf{0.001}$_{0.004}$  \\
        &CTGAN$^\dagger$ & 0.013$_{0.001}$ & 0.020$_{0.015}$ & 0.005$_{0.003}$ & \textbf{0.003}$_{0.002}$ & 0.004$_{0.000}$  \\
         &ADS-GAN$^\dagger$ & \textbf{0.006}$_{0.002}$ & 0.009$_{0.005}$ & 0.005$_{0.001}$ & 0.004$_{0.004}$ & 0.010$_{0.015}$  \\
        \cmidrule{2-7}
        
         & TVAE  & 0.008$_{0.002}$ & 0.009$_{0.000}$ & \textbf{0.004}$_{0.005}$ & 0.005$_{0.002}$ & 0.003$_{0.004}$ \\
          &TabDDPM  & \textbf{0.006}$_{0.005}$ & 0.009$_{0.004}$ & 0.014$_{0.002}$ & 0.004$_{0.000}$ & \textbf{0.001}$_{0.003}$ \\
          &CTGAN  & 0.013$_{0.004}$ & 0.007$_{0.002}$ & 0.005$_{0.005}$ & 0.004$_{0.002}$ & 0.003$_{0.001}$ \\
         &ADS-GAN  & 0.006$_{0.003}$ & \textbf{0.006}$_{0.002}$ & 0.005$_{0.004}$ & 0.005$_{0.005}$ & 0.005$_{0.000}$ \\
         
         \cmidrule{2-7}
         & TVAE $(u)$ & 0.011$_{0.002}$ & 0.009$_{0.005}$ & 0.007$_{0.001}$ & 0.007$_{0.004}$ & 0.003$_{0.000}$ \\
          &TabDDPM $(u)$ & \textbf{0.006}$_{0.001}$ & 0.007$_{0.004}$ & 0.006$_{0.002}$ & 0.005$_{0.005}$ & 0.002$_{0.000}$ \\
          &CTGAN $(u)$ & 0.007$_{0.005}$ & 0.012$_{0.002}$ & 0.005$_{0.004}$ & 0.008$_{0.001}$ & 0.005$_{0.003}$ \\
         &ADS-GAN $(u)$ & \textbf{0.006}$_{0.002}$ & 0.007$_{0.000}$ & 0.007$_{0.005}$ & 0.005$_{0.002}$ & 0.005$_{0.004}$ \\

         \midrule

        \multirow{13}{*}{WS distance ($\downarrow$)} & SurvivalGAN & 0.112$_{0.015}$ & 0.039$_{0.002}$ & 0.043$_{0.004}$ & 0.019$_{0.005}$ & 0.052$_{0.000}$\\
         & TVAE$^\dagger$ & \textbf{0.061}$_{0.004}$ & \textbf{0.028}$_{0.005}$ & \textbf{0.032}$_{0.002}$ & 0.013$_{0.000}$ & \textbf{0.016}$_{0.005}$ \\
        &TabDDPM$^\dagger$ & 0.159$_{0.025}$ & 0.089$_{0.004}$ & 0.308$_{0.025}$ & 0.056$_{0.002}$ & 0.028$_{0.005}$  \\
        &CTGAN$^\dagger$ & 0.095$_{0.002}$ & 0.133$_{0.015}$ & 0.034$_{0.005}$ & 0.013$_{0.004}$ & 0.019$_{0.000}$  \\
         &ADS-GAN$^\dagger$ & 0.082$_{0.005}$ & 0.037$_{0.002}$ & 0.036$_{0.004}$ & \textbf{0.011}$_{0.005}$ & 0.018$_{0.004}$  \\
         
            \cmidrule{2-7}
            
         & TVAE  & 0.063$_{0.004}$ & 0.037$_{0.005}$ & 0.032$_{0.002}$ & \textbf{0.011}$_{0.004}$ & \textbf{0.016}$_{0.002}$ \\
          &TabDDPM  & 0.160$_{0.005}$ & 0.104$_{0.002}$ & 0.309$_{0.004}$ & 0.037$_{0.005}$ & 0.028$_{0.000}$ \\
          &CTGAN  & 0.097$_{0.002}$ & 0.037$_{0.004}$ & 0.035$_{0.005}$ & 0.014$_{0.002}$ & 0.018$_{0.004}$ \\
         &ADS-GAN  & 0.079$_{0.005}$ & 0.030$_{0.002}$ & 0.034$_{0.000}$ & \textbf{0.011}$_{0.005}$ & 0.017$_{0.002}$ \\
         
         \cmidrule{2-7}
         
          &TVAE $(u)$ & 0.075$_{0.004}$ & 0.031$_{0.005}$ & 0.037$_{0.002}$ & 0.013$_{0.004}$ & 0.017$_{0.005}$ \\
          &TabDDPM $(u)$ & 0.079$_{0.005}$ & 0.031$_{0.002}$ & 0.049$_{0.004}$ & 0.015$_{0.000}$ & \textbf{0.016}$_{0.005}$ \\
          &CTGAN $(u)$ & 0.069$_{0.000}$ & 0.041$_{0.004}$ & 0.036$_{0.005}$ & 0.017$_{0.002}$ & 0.021$_{0.004}$ \\
         &ADS-GAN $(u)$ & 0.065$_{0.005}$ & 0.035$_{0.000}$ & 0.038$_{0.004}$ & 0.013$_{0.005}$ & 0.017$_{0.002}$ \\
         
       \midrule
       
         \multirow{14}{*}{C-Index ($\uparrow$)} & SurvivalGAN & 0.735$_{0.005}$ & 0.625$_{0.000}$ & 0.602$_{0.004}$ & 0.668$_{0.005}$ & 0.870$_{0.002}$\\
         & TVAE$^\dagger$ & 0.737$_{0.004}$ & 0.612$_{0.005}$ & 0.583$_{0.000}$ & 0.672$_{0.001}$ & 0.872$_{0.005}$ \\
        &TabDDPM$^\dagger$ & 0.660$_{0.075}$ & 0.589$_{0.015}$ & 0.536$_{0.005}$ & 0.663$_{0.002}$ & 0.876$_{0.004}$  \\
        &CTGAN$^\dagger$ & 0.746$_{0.005}$ & 0.628$_{0.015}$ & 0.577$_{0.004}$ & 0.665$_{0.015}$ & 0.874$_{0.002}$  \\
         &ADS-GAN$^\dagger$ & \textbf{0.797}$_{0.015}$ & \textbf{0.655}$_{0.005}$ & 0.623$_{0.002}$ & 0.684$_{0.004}$ & \textbf{0.880}$_{0.005}$  \\
                \cmidrule{2-7}

         & TVAE  & 0.783$_{0.025}$ & 0.630$_{0.005}$ & 0.602$_{0.004}$ & 0.672$_{0.002}$ & 0.868$_{0.005}$ \\
          &TabDDPM  & 0.670$_{0.015}$ & 0.603$_{0.004}$ & 0.530$_{0.005}$ & 0.659$_{0.002}$ & 0.875$_{0.004}$ \\
          &CTGAN  & 0.760$_{0.005}$ & 0.629$_{0.004}$ & 0.602$_{0.002}$ & 0.668$_{0.005}$ & 0.874$_{0.000}$ \\
         &ADS-GAN  & 0.785$_{0.025}$ & 0.652$_{0.005}$ & \textbf{0.626}$_{0.004}$ & 0.682$_{0.002}$ & \textbf{0.880}$_{0.005}$ \\
         
         \cmidrule{2-7}

          & TVAE $(u)$ & 0.735$_{0.004}$ & 0.646$_{0.002}$ & 0.604$_{0.005}$ & 0.671$_{0.004}$ & 0.878$_{0.005}$\\
        &TabDDPM $(u)$& 0.759$_{0.005}$ & 0.649$_{0.004}$ & 0.625$_{0.002}$ & 0.679$_{0.005}$ & 0.879$_{0.004}$\\
        &CTGAN $(u)$& 0.779$_{0.002}$ & 0.647$_{0.005}$ & 0.606$_{0.004}$ & 0.679$_{0.002}$ & 0.878$_{0.005}$\\
         &ADS-GAN $(u)$ & 0.776$_{0.004}$ & 0.636$_{0.005}$ & 0.601$_{0.002}$ & 0.663$_{0.004}$ & 0.878$_{0.002}$\\
         \midrule
         
        \multirow{14}{*}{Brier Score ($\downarrow$)} &  SurvivalGAN & 0.068$_{0.005}$ & 0.205$_{0.004}$ & 0.202$_{0.002}$ & 0.212$_{0.005}$ & 0.096$_{0.004}$\\
         & TVAE$^\dagger$ & \textbf{0.059}$_{0.004}$ & 0.199$_{0.005}$ & 0.207$_{0.004}$ & 0.214$_{0.002}$ & 0.095$_{0.005}$ \\
        &TabDDPM$^\dagger$ & 0.063$_{0.005}$ & 0.212$_{0.004}$ & 0.217$_{0.002}$ & 0.215$_{0.005}$ & 0.096$_{0.004}$  \\
        &CTGAN$^\dagger$ & 0.061$_{0.004}$ & 0.199$_{0.005}$ & 0.205$_{0.004}$ & 0.215$_{0.015}$ & 0.089$_{0.005}$  \\
         &ADS-GAN$^\dagger$ & \textbf{0.059}$_{0.005}$ & \textbf{0.197}$_{0.004}$ & 0.198$_{0.002}$ & 0.213$_{0.005}$ & \textbf{0.084}$_{0.004}$  \\
         
                \cmidrule{2-7}
                
         & TVAE  & 0.060$_{0.004}$ & \textbf{0.197}$_{0.005}$ & 0.208$_{0.004}$ & 0.211$_{0.005}$ & 0.091$_{0.002}$ \\
          &TabDDPM  & 0.061$_{0.005}$ & 0.209$_{0.004}$ & 0.218$_{0.005}$ & 0.210$_{0.004}$ & 0.091$_{0.002}$ \\
          &CTGAN  & 0.060$_{0.004}$ & 0.199$_{0.005}$ & 0.204$_{0.002}$ & 0.211$_{0.004}$ & 0.089$_{0.005}$ \\
         &ADS-GAN  & 0.061$_{0.005}$ & 0.195$_{0.004}$ & \textbf{0.198}$_{0.002}$ & 0.214$_{0.005}$ & 0.085$_{0.004}$ \\

         \cmidrule{2-7}
         
        &TVAE $(u)$& 0.061$_{0.004}$ & 0.204$_{0.005}$ & 0.206$_{0.004}$ & 0.210$_{0.002}$ & 0.093$_{0.005}$ \\
        &TabDDPM $(u)$& 0.060$_{0.005}$ & 0.200$_{0.004}$ & 0.199$_{0.002}$ & 0.207$_{0.005}$ & 0.087$_{0.004}$ \\
        &CTGAN $(u)$ & 0.064$_{0.004}$ & 0.202$_{0.005}$ & 0.203$_{0.004}$ & 0.210$_{0.002}$ & 0.086$_{0.005}$ \\
        &ADSGAN $(u)$ & 0.061$_{0.005}$ & 0.207$_{0.004}$ & 0.201$_{0.002}$ & 0.208$_{0.005}$ & 0.088$_{0.004}$ \\
       \bottomrule
        \end{tabular}
    }
    \label{tab:full baselines}
\end{table}

\begin{table}[h]
    \small
    \caption{\small Event-time distribution quality metrics. Models conditioned on empirical $t$ and $e$ are highlighted ($\dagger$) and $u$ refers to unconditional models. Subscripts are standard deviations for 5 repetitions.}
    \vspace{1mm}
    \adjustbox{width=\textwidth}{
        \begin{tabular}{ccccccc}
        \toprule
        \textbf{Metric}&\textbf{Method} & \textbf{AIDS} & \textbf{METABRIC} & \textbf{SUPPORT} & \textbf{GBSG}& \textbf{FLCHAIN}\\
        \hline
        \multirow{13}{*}{Optimism (\small{$\rightarrow$ 0})} & SurvivalGAN & 0.021$_{0.005}$ & 0.011$_{0.002}$ & 0.016$_{0.004}$ & 0.006$_{0.003}$ & 0.134$_{0.005}$ \\
& TVAE$^\dagger$ & \textbf{0.000}$_{0.001}$ & \textbf{0.000}$_{0.002}$ & \textbf{0.000}$_{0.000}$ & \textbf{0.003}$_{0.001}$ & \textbf{0.001}$_{0.002}$ \\
& TabDDPM$^\dagger$ & \textbf{0.000}$_{0.001}$ & \textbf{0.000}$_{0.002}$ & \textbf{0.000}$_{0.000}$ & \textbf{0.003}$_{0.001}$ & \textbf{0.001}$_{0.002}$ \\
& CTGAN$^\dagger$ & \textbf{0.000}$_{0.001}$ & \textbf{0.000}$_{0.002}$ & \textbf{0.000}$_{0.000}$ & \textbf{0.003}$_{0.001}$ & \textbf{0.001}$_{0.002}$\\
& ADSGAN$^\dagger$ & \textbf{0.000}$_{0.001}$ & \textbf{0.000}$_{0.002}$ & \textbf{0.000}$_{0.000}$ & \textbf{0.003}$_{0.001}$ & \textbf{0.001}$_{0.002}$ \\

                \cmidrule{2-7}
                
         & TVAE  & 0.001$_{0.001}$ & 0.001$_{0.001}$ & -0.001$_{0.001}$ & -0.004$_{0.000}$ & -0.003$_{0.001}$ \\
          &TabDDPM  & 0.001$_{0.001}$ & 0.001$_{0.001}$ & -0.001$_{0.001}$ & -0.004$_{0.000}$ & -0.003$_{0.001}$ \\
          &CTGAN  & 0.001$_{0.001}$ & 0.001$_{0.001}$ & -0.001$_{0.001}$ & -0.004$_{0.000}$ & -0.003$_{0.001}$ \\
         &ADS-GAN  & 0.001$_{0.001}$ & 0.001$_{0.001}$ & -0.001$_{0.001}$ & -0.004$_{0.000}$ & -0.003$_{0.001}$ \\

         \cmidrule{2-7}
         
& TVAE $(u)$ & 0.023$_{0.003}$ & -0.003$_{0.004}$ & -0.014$_{0.002}$ & 0.004$_{0.005}$ & 0.022$_{0.003}$ \\
& TabDDPM $(u)$ & 0.021$_{0.004}$ & 0.001$_{0.005}$ & 0.001$_{0.003}$ & 0.026$_{0.002}$ & 0.005$_{0.004}$ \\
& CTGAN $(u)$& -0.005$_{0.005}$ & 0.017$_{0.002}$ & -0.038$_{0.004}$ & 0.060$_{0.003}$ & -0.037$_{0.005}$ \\
& ADSGAN $(u)$ & 0.002$_{0.002}$ & -0.033$_{0.003}$ & -0.007$_{0.005}$ & 0.010$_{0.004}$ & 0.005$_{0.002}$ \\

\midrule

\multirow{13}{*}{Short Sightedness (\small{$\rightarrow$ 0})} & SurvivalGAN & 0.007$_{0.003}$ & 0.124$_{0.004}$ & 0.020$_{0.002}$ & 0.019$_{0.005}$ & 0.005$_{0.003}$ \\
& TVAE$^\dagger$ &\textbf{0.001}$_{0.000}$ & \textbf{0.000}$_{0.000}$ & \textbf{0.000}$_{0.001}$ & \textbf{0.010}$_{0.012}$ & \textbf{0.002}$_{0.001}$ \\
& TabDDPM$^\dagger$ & \textbf{0.001}$_{0.000}$ & \textbf{0.000}$_{0.000}$ & \textbf{0.000}$_{0.001}$ & \textbf{0.010}$_{0.012}$ & \textbf{0.002}$_{0.001}$ \\
& CTGAN$^\dagger$ & \textbf{0.001}$_{0.000}$ & \textbf{0.000}$_{0.000}$ & \textbf{0.000}$_{0.001}$ & \textbf{0.010}$_{0.012}$ & \textbf{0.002}$_{0.001}$ \\
& ADS-GAN$^\dagger$ & \textbf{0.001}$_{0.000}$ & \textbf{0.000}$_{0.000}$ & \textbf{0.000}$_{0.001}$ & \textbf{0.010}$_{0.012}$ & \textbf{0.002}$_{0.001}$ \\
                \cmidrule{2-7}
                
         & TVAE  & \textbf{0.001}$_{0.001}$ & \textbf{0.000}$_{0.001}$ & \textbf{0.000}$_{0.002}$ & -0.012$_{0.001}$ & -0.002$_{0.002}$ \\
          &TabDDPM  & \textbf{0.001}$_{0.001}$ & \textbf{0.000}$_{0.001}$ & \textbf{0.000}$_{0.002}$ & -0.012$_{0.001}$ & -0.002$_{0.002}$ \\
          &CTGAN  & \textbf{0.001}$_{0.001}$ & \textbf{0.000}$_{0.001}$ & \textbf{0.000}$_{0.002}$ & -0.012$_{0.001}$ & -0.002$_{0.002}$ \\
         &ADS-GAN  & \textbf{0.001}$_{0.001}$ & \textbf{0.000}$_{0.001}$ & \textbf{0.000}$_{0.002}$ & -0.012$_{0.001}$ & -0.002$_{0.002}$ \\

         \cmidrule{2-7}
         
& TVAE $(u)$ & 0.058$_{0.004}$ & 0.148$_{0.005}$ & 0.002$_{0.002}$ & 0.017$_{0.003}$ & 0.018$_{0.004}$ \\
& TabDDPM $(u)$& 0.002$_{0.005}$ & 0.000$_{0.002}$ & 0.002$_{0.003}$ & 0.015$_{0.004}$ & 0.003$_{0.005}$ \\
& CTGAN $(u)$& 0.071$_{0.002}$ & 0.056$_{0.003}$ & 0.010$_{0.004}$ & 0.019$_{0.005}$ & 0.017$_{0.002}$ \\
& ADSGAN $(u)$& 0.040$_{0.003}$ & 0.188$_{0.004}$ & 0.002$_{0.005}$ & 0.014$_{0.002}$ & 0.006$_{0.003}$ \\

\midrule

\multirow{13}{*}{KM Divergence (\small{$\downarrow$})}& SurvivalGAN & 0.021$_{0.004}$ & 0.082$_{0.005}$ & 0.064$_{0.002}$ & 0.049$_{0.003}$ & 0.134$_{0.004}$ \\
& TVAE$^\dagger$ & \textbf{0.002}$_{0.000}$ & \textbf{0.008}$_{0.001}$ & \textbf{0.002}$_{0.000}$ & \textbf{0.005}$_{0.001}$ & \textbf{0.002}$_{0.000}$ \\
& TabDDPM$^\dagger$ &\textbf{0.002}$_{0.000}$ & \textbf{0.008}$_{0.001}$ & \textbf{0.002}$_{0.000}$ & \textbf{0.005}$_{0.001}$ & \textbf{0.002}$_{0.000}$ \\
& CTGAN$^\dagger$ & \textbf{0.002}$_{0.000}$ & \textbf{0.008}$_{0.001}$ & \textbf{0.002}$_{0.000}$ & \textbf{0.005}$_{0.001}$ & \textbf{0.002}$_{0.000}$\\
& ADS-GAN$^\dagger$ & \textbf{0.002}$_{0.000}$ & \textbf{0.008}$_{0.001}$ & \textbf{0.002}$_{0.000}$ & \textbf{0.005}$_{0.001}$ & \textbf{0.002}$_{0.000}$ \\

                \cmidrule{2-7}
                
         & TVAE  & 0.003$_{0.001}$ & 0.013$_{0.002}$ & 0.006$_{0.001}$ & 0.007$_{0.001}$ & 0.005$_{0.001}$ \\
          &TabDDPM  & 0.003$_{0.001}$ & 0.013$_{0.002}$ & 0.006$_{0.001}$ & 0.007$_{0.001}$ & 0.005$_{0.001}$ \\
          &CTGAN  & 0.003$_{0.001}$ & 0.013$_{0.002}$ & 0.006$_{0.001}$ & 0.007$_{0.001}$ & 0.005$_{0.001}$ \\
         &ADS-GAN  & 0.003$_{0.001}$ & 0.013$_{0.002}$ & 0.006$_{0.001}$ & 0.007$_{0.001}$ & 0.005$_{0.001}$ \\

         \cmidrule{2-7}
         
($\downarrow$)& TVAE $(u)$ & 0.031$_{0.005}$ & 0.042$_{0.002}$ & 0.025$_{0.003}$ & 0.027$_{0.004}$ & 0.031$_{0.005}$ \\
& TabDDPM $(u)$& 0.021$_{0.002}$ & 0.019$_{0.003}$ & 0.011$_{0.004}$ & 0.026$_{0.005}$ & 0.007$_{0.002}$ \\
& CTGAN $(u)$& 0.015$_{0.003}$ & 0.028$_{0.004}$ & 0.038$_{0.005}$ & 0.061$_{0.002}$ & 0.037$_{0.003}$ \\
& ADSGAN $(u)$& 0.016$_{0.004}$ & 0.039$_{0.005}$ & 0.020$_{0.002}$ & 0.030$_{0.003}$ & 0.012$_{0.004}$\\
         \bottomrule
        \end{tabular}
    }
    \label{tab:event metrics}
\end{table}

\begin{table*}[h]
    \centering
    \caption{\small Quality, downstream and event-time metrics. Performance reported for our best model between ADS-GAN, TVAE, CTGAN and TabDDPM from Table \ref{tab:full baselines} and Table \ref{tab:event metrics} for both settings. Models conditioned on empirical $t$ and $e$ are highlighted ($\dagger$). Subscripts are standard deviations for 5 repetitions.}
    \vspace{1mm}
    \adjustbox{width=0.98\textwidth}{
        \begin{tabular}{c c c c c c c}
        \toprule
        \textbf{Metric}&\textbf{Method} & \textbf{AIDS} & \textbf{METABRIC} & \textbf{SUPPORT} & \textbf{GBSG}& \textbf{FLCHAIN}\\
        \hline
        
         \multirow{2}{*}{JS distance ($\downarrow$)}&Ours & \textbf{0.006}$_{0.005}$ & \textbf{0.006}$_{0.002}$ & \textbf{0.004}$_{0.005}$ &0.004$_{0.002}$ &\textbf{0.001}$_{0.003}$ \\
          & Ours$^\dagger$ & \textbf{0.006}$_{0.002}$ & 0.007$_{0.001}$ &\textbf{0.004}$_{0.000}$ & \textbf{0.003}$_{0.002}$ &\textbf{0.001}$_{0.004}$ \\
        
         \cmidrule{2-7}
        \multirow{2}{*}{WS distance ($\downarrow$)}&Ours & 0.063$_{0.004}$ & 0.030$_{0.002}$ & \textbf{0.032}$_{0.002}$ &\textbf{0.011}$_{0.004}$ &\textbf{0.016}$_{0.002}$ \\
        & Ours$^\dagger$ & \textbf{0.061}$_{0.004}$ & \textbf{0.028}$_{0.005}$ &\textbf{0.032}$_{0.002}$ & \textbf{0.011}$_{0.005}$ &\textbf{0.016}$_{0.005}$ \\

       \midrule
       
         \multirow{2}{*}{C-Index ($\uparrow$)} &Ours & 0.785$_{0.025}$ & 0.652$_{0.005}$ & \textbf{0.626}$_{0.004}$ &0.682$_{0.002}$ &\textbf{0.880}$_{0.005}$ \\
         & Ours$^\dagger$ & \textbf{0.797}$_{0.015}$ & \textbf{0.655}$_{0.005}$ &0.623$_{0.002}$ & \textbf{0.684}$_{0.004}$ &\textbf{0.880}$_{0.005}$ \\

         \cmidrule{2-7}
         
        \multirow{2}{*}{Brier Score ($\downarrow$)}&Ours & 0.060$_{0.004}$ & \textbf{0.197}$_{0.005}$ & \textbf{0.198}$_{0.002}$ &0.210$_{0.004}$ &0.085$_{0.004}$ \\
        & Ours$^\dagger$ & \textbf{0.059}$_{0.005}$ & \textbf{0.197}$_{0.004}$ &\textbf{0.198}$_{0.002}$ & 0.213$_{0.005}$ &\textbf{0.084}$_{0.004}$  \\

         \midrule
         
       \multirow{2}{*}{Optimism ($\rightarrow 0$)}&Ours & 0.001$_{0.001}$ & 0.001$_{0.001}$ & -0.001$_{0.001}$ &-0.004$_{0.000}$ &-0.003$_{0.001}$ \\
       & Ours$^\dagger$ & \textbf{0.000}$_{0.001}$ & \textbf{0.000}$_{0.002}$ &\textbf{0.000}$_{0.000}$ & \textbf{0.003}$_{0.001}$ &\textbf{0.001}$_{0.002}$ \\
    
         \cmidrule{2-7}

        \multirow{2}{*}{Short Sightedness ($\rightarrow 0$)}&Ours & \textbf{0.001}$_{0.001}$ & \textbf{0.000}$_{0.001}$ & \textbf{0.000}$_{0.002}$ &-0.012$_{0.001}$ &\textbf{-0.002}$_{0.002}$ \\
        & Ours$^\dagger$ & \textbf{0.001}$_{0.000}$ & \textbf{0.000}$_{0.000}$ &\textbf{0.000}$_{0.001}$ & \textbf{0.010}$_{0.012}$ &\textbf{0.002}$_{0.001}$ \\

         \cmidrule{2-7}
         
      \multirow{2}{*}{KM Divergence ($\downarrow$)}&Ours & 0.003$_{0.001}$ & 0.013$_{0.002}$ & 0.006$_{0.001}$ &0.007$_{0.001}$ &0.005$_{0.001}$ \\
      & Ours$^\dagger$ &\textbf{0.002}$_{0.000}$ & \textbf{0.008}$_{0.001}$ &\textbf{0.002}$_{0.000}$ & \textbf{0.005}$_{0.001}$ &\textbf{0.002}$_{0.000}$ \\
         \bottomrule
         
        \end{tabular}
    }
    \label{tab:ablation}
\end{table*}

\begin{table}[t]
    \centering
    \small
    \caption{Median value of Distance of closest record from the original. Models conditioned on empirical $t$ and $e$ are highlighted ($\dagger$) and $u$ refers to unconditional models. Subscripts are standard deviations for 5 repetitions. The highest (best) values are in bold and the least (worst) values are underlined.}
    {
        \begin{tabular}{ccccccc}
        \hline
        \textbf{Metric}&\textbf{Method} & \textbf{AIDS} & \textbf{METABRIC} & \textbf{SUPPORT} & \textbf{GBSG}& \textbf{FLCHAIN}\\
        \hline
         \multirow{14}{*}{} & SurvivalGAN & 1.035$_{0.001}$ & 0.969$_{0.004}$ & 1.589$_{0.002}$ & 0.500$_{0.003}$ & \textbf{0.796}$_{0.004}$\\
         & TVAE$^\dagger$ & 0.883$_{0.002}$ & 0.877$_{0.003}$ & 1.511$_{0.002}$ & 0.476$_{0.002}$ & 0.642$_{0.003}$ \\
        & TabDDPM$^\dagger$ & 1.172$_{0.031}$ & 0.908$_{0.015}$ & 1.612$_{0.035}$ & 0.519$_{0.004}$ & 0.572$_{0.013}$  \\
        & CTGAN$^\dagger$ & 0.918$_{0.013}$ & 1.043$_{0.004}$ & 1.594$_{0.001}$ & \textbf{0.524}$_{0.003}$ & 0.695$_{0.004}$ \\
         Median & ADS-GAN$^\dagger$ & 1.133$_{0.170}$ & 0.992$_{0.006}$ & 1.691$_{0.004}$ & 0.519$_{0.003}$ & 0.667$_{0.014}$  \\
         DCR & SMOTE & \underline{0.388}$_{0.002}$ & \underline{0.698}$_{0.001}$ & \underline{0.958}$_{0.003}$ & \underline{0.290}$_{0.005}$ & \underline{0.381}$_{0.006}$  \\
         \cmidrule{2-7}
         & TVAE & 0.992$_{0.002}$ & 0.881$_{0.001}$ & 1.523$_{0.004}$ & 0.481$_{0.002}$ & 0.648$_{0.003}$ \\
        & TabDDPM & \textbf{1.184}$_{0.034}$ & 0.915$_{0.015}$ & 1.625$_{0.035}$ & 0.522$_{0.004}$ & 0.578$_{0.013}$  \\
        & CTGAN & 1.007$_{0.012}$ & 1.052$_{0.001}$ & 1.602$_{0.003}$ & 0.521$_{0.002}$ & 0.689$_{0.004}$ \\
         & ADS-GAN & 1.142$_{0.175}$ & 0.998$_{0.003}$ & \textbf{1.698}$_{0.007}$ & 0.523$_{0.003}$ & 0.672$_{0.014}$  \\
        \cmidrule{2-7}
        & TVAE $(u)$ & 1.044$_{0.023}$ & 0.813$_{0.004}$ & 1.405$_{0.003}$ & 0.432$_{0.003}$ & 0.553$_{0.001}$ \\
        & TabDDPM $(u)$ & 1.020$_{0.013}$ & \textbf{1.087}$_{0.004}$ & 1.611$_{0.006}$ & 0.477$_{0.004}$ & 0.567$_{0.003}$ \\
        & CTGAN $(u)$ & 1.112$_{0.014}$ & 1.001$_{0.003}$ & 1.586$_{0.001}$ & 0.515$_{0.003}$ & 0.641$_{0.004}$ \\
         & ADS-GAN $(u)$ & 1.158$_{0.011}$ & 0.945$_{0.002}$ & 1.666$_{0.003}$ & 0.475$_{0.004}$ & 0.533$_{0.001}$  \\
         \bottomrule
        \end{tabular}
    }
    \label{tab:privacy median full}
\end{table}

\begin{table}[b]
   \centering
   \small
   \caption{Minimum value of Distance of closest record from the original. Models conditioned on empirical $t$ and $e$ are highlighted ($\dagger$) and $u$ refers to unconditional models. Subscripts are standard deviations for 5 repetitions. The highest (best) values are in bold and the least (worst) values are underlined.}
   {
       \begin{tabular}{ccccccc}
       \hline
       \textbf{Metric}&\textbf{Method} & \textbf{AIDS} & \textbf{METABRIC} & \textbf{SUPPORT} & \textbf{GBSG}& \textbf{FLCHAIN}\\
       \hline
        \multirow{14}{*}{} & SurvivalGAN & 0.048$_{0.003}$ & 0.172$_{0.004}$ & 0.326$_{0.003}$ & 0.062$_{0.002}$ & 0.057$_{0.003}$\\
        & TVAE$^\dagger$ & 0.077$_{0.034}$ & 0.202$_{0.025}$ & 0.370$_{0.023}$ & 0.033$_{0.004}$ & 0.026$_{0.003}$ \\
       & TabDDPM$^\dagger$ & 0.095$_{0.003}$ & 0.193$_{0.055}$ & 0.403$_{0.013}$ & 0.065$_{0.004}$ & 0.037$_{0.002}$  \\
       & CTGAN$^\dagger$ & \textbf{0.139}$_{0.015}$ & \textbf{0.215}$_{0.014}$ & 0.321$_{0.015}$ & 0.045$_{0.013}$ & 0.054$_{0.014}$ \\
        Minimum & ADS-GAN$^\dagger$ & 0.102$_{0.013}$ & 0.185$_{0.043}$ & 0.391$_{0.015}$ & 0.053$_{0.015}$ & 0.066$_{0.025}$  \\
        DCR & SMOTE & \underline{0.000}$_{0.000}$ & \underline{0.000}$_{0.001}$ & \underline{0.000}$_{0.002}$ & \underline{0.000}$_{0.001}$ & \underline{0.000}$_{0.002}$  \\
        \cmidrule{2-7}
        & TVAE & 0.081$_{0.014}$ & 0.208$_{0.011}$ & 0.378$_{0.024}$ & 0.035$_{0.003}$ & 0.028$_{0.004}$ \\
       & TabDDPM & 0.098$_{0.004}$ & 0.198$_{0.040}$ & 0.412$_{0.015}$ & \textbf{0.068}$_{0.003}$ & 0.039$_{0.003}$  \\
       & CTGAN & 0.135$_{0.011}$ & 0.211$_{0.016}$ & 0.325$_{0.017}$ & 0.048$_{0.015}$ & 0.058$_{0.016}$ \\
        & ADS-GAN & 0.106$_{0.015}$ & 0.189$_{0.045}$ & 0.395$_{0.017}$ & 0.056$_{0.017}$ & \textbf{0.069}$_{0.021}$  \\
       \cmidrule{2-7}
       & TVAE $(u)$ & 0.083$_{0.013}$ & 0.154$_{0.037}$ & 0.171$_{0.014}$ & 0.031$_{0.003}$ & 0.028$_{0.004}$ \\
       & TabDDPM $(u)$ & 0.090$_{0.035}$ & 0.213$_{0.004}$ & 0.337$_{0.013}$ & 0.054$_{0.004}$ & 0.033$_{0.003}$ \\
       & CTGAN $(u)$ & 0.109$_{0.025}$ & 0.194$_{0.023}$ & 0.316$_{0.025}$ & 0.046$_{0.015}$ & 0.024$_{0.004}$ \\
        & ADS-GAN $(u)$ & 0.062$_{0.037}$ & 0.205$_{0.035}$ & \textbf{0.429}$_{0.004}$ & 0.050$_{0.004}$ & 0.055$_{0.003}$  \\
        \bottomrule
       \end{tabular}
   }
   \label{tab:privacy min full}
\end{table}

\end{document}